# Physics-Informed Deep Learning to Reduce the Bias in Joint Prediction of Nitrogen Oxides

Lianfa Li[1,2]*, Roxana Khalili[1], Frederick Lurmann[3], Nathan Pavlovic[3], Jun Wu[4], Yan Xu[1], Yisi Liu[1], Karl O'Sharkey[1], Beate Ritz[5], Luke Oman[6], Meredith Franklin[1,7], Theresa Bastain[1], Shohreh F. Farzan[1], Carrie Breton[1], Rima Habre[1,8]*

[1]Department of Population and Public Health Sciences, University of Southern California, Los Angeles, CA, USA.

[2]State Key Laboratory of Resources and Environmental Information System, Institute of Geographical Sciences and Natural Resources, Chinese Academy of Sciences, Beijing, China.

[3]Sonoma Technology, Inc., Petaluma, CA, USA.

[4]Program in Public Health, Susan and Henry Samueli College of Health Sciences, University of California, Irvine, CA, USA.

[5]Departments of Epidemiology and Environmental Health, Fielding School of Public Health, University of California, Los Angeles, CA, USA.

[6]Goddard Space Flight Center, National Aeronautics and Space Administration, Greenbelt, MD, USA.

[7] Department of Statistical Sciences, University of Toronto, Toronto, Ontario Canada.

[8] Spatial Sciences Institute, University of Southern California, Los Angeles, CA, USA.

*Corresponding author. Email: lianfali@usc.edu (L.L.); habre@usc.edu (R.H.).

**Abstract:** Atmospheric nitrogen oxides ($NO_x$) primarily from fuel combustion have recognized acute and chronic health and environmental effects. Machine learning (ML) methods have significantly enhanced our capacity to predict $NO_x$ concentrations at ground-level with high spatiotemporal resolution but may suffer from high estimation bias since they lack physical and chemical knowledge about air pollution dynamics. Chemical transport models (CTMs) leverage this knowledge; however, accurate predictions of ground-level concentrations typically necessitate extensive post-calibration. Here, we present a physics-informed deep learning framework that encodes advection-diffusion mechanisms and fluid dynamics constraints to jointly predict $NO_2$ and $NO_x$ and reduce ML model bias by 21-42%. Our approach captures fine-scale transport of $NO_2$ and $NO_x$, generates robust spatial extrapolation, and provides explicit uncertainty estimation. The framework fuses knowledge-driven physicochemical principles of CTMs with the predictive power of ML for air quality exposure, health, and policy applications. Our approach offers significant improvements over purely data-driven ML methods and has unprecedented bias reduction in joint $NO_2$ and $NO_x$ prediction.

Nitrogen oxides ($NO_x$) - a mixture of nitrogen dioxide ($NO_2$), nitric oxide (NO) and other nitrogen species are emitted primarily from anthropogenic combustion sources such as motor vehicles and also formed through secondary pathways. Short-term health effects associated with $NO_x$ include airway irritation and other acute respiratory diseases (*1-3*). Long-term exposures to $NO_x$ can lead to the development of asthma, increase in the susceptibility to respiratory infections, and mortality (*4, 5*). As such, $NO_2$ is a criteria pollutant that is regulated under the U.S. Clean Air Act and many other countries worldwide given its documented adverse effects on public health. NO and $NO_2$ are reactive air pollutants that are affected by complex atmospheric physical processes and chemical reactions from emissions and transport to dry and wet deposition (Fig 1) (*6-10*). The majority of freshly emitted NO converts to $NO_2$ in the presence of ozone and volatile organic compounds (VOCs) on timescales from seconds to hours (Fig. 1A). $NO_2$ has a longer atmospheric lifetime than NO, particularly in winter when cool, dry conditions in the troposphere slow down photochemical reactions, allowing pollutants to accumulate or transport across regions below or above temperature inversions (Fig. 1B). $NO_2$ may react with other chemicals in the air to form nitric acid, particulate nitrate ($PM-NO_3$) and peroxyacyl nitrates (PANs), and under the presence of sunlight $NO_2$ can convert back to NO and produce ozone ($O_3$) (*11, 12*) (Fig. 1C). $NO_2$ also interacts with water, oxygen and other chemicals to form acid rain that harms sensitive ecosystems such as lakes and forests (Fig. 1D). Under unfavorable meteorological conditions, $NO_2$ and its secondary derived air pollutants are key contributors to photochemical smog that is harmful to health and the environment (*13, 14*). As such, accurately predicting $NO_x$ concentrations and the $NO_2$ contribution to $NO_x$ has important societal benefits for health and environment protection.

$NO_x$ concentrations have significant variability at small spatial scales and high gradients away from roadways (*15-17*), making its prediction at local scales challenging. As an important measure of $NO_x$ prediction performance, estimation bias is defined as the error between a model's predicted values and measured observations, describing how well the model can reproduce real data (*18*). Some of the major sources of estimation bias in $NO_x$ models include limited or outdated emissions data, incorrect characterization of transport and chemical transformations, and insufficient information to capture pollutant deposition (Fig. 1E). In addition to meteorology, traffic density and land-use variables have been used as variables in models to represent $NO_x$ emissions and sinks (*19*); however, this is not sufficient to capture $NO_x$ dynamics. Chemical transport models (CTMs) and dispersion models use mathematical representations of chemical and/or physical processes to simulate transport and transformation of air pollutants such as $NO_x$, including emissions, plume rise, transport diffusion/dispersion, photochemistry, deposition and secondary formation (*20*). These deterministic CTMs are often applied at coarse spatial scales due to their high computational cost limiting their potential for meeting the increasing demand for fine-grained predictions. Likewise, physical dispersion models may have fine spatial scale but generally do not capture chemical transformations. Purely statistical machine learning (ML) models, such as random forest, gradient boosting and neural networks, rely solely on data to reproduce statistical relationships between features (covariates) and $NO_x$ (*21*), or are combined with physical models (*22*) to speed up physiochemical integral computing. However, when a ML model is only trained to reproduce statistical relationships or patterns in the data, it does not guarantee that its predictions preserve the physical laws of air pollution fluid dynamics – a problem which is further enhanced when generalizing to areas without training data - which leads to high estimation bias (*23*). Furthermore, commonly utilized model cross-validation techniques (i.e., where training data is randomly split across space and time and a subset held out for testing) do not provide a true generalization of the trained model at new spatial locations, potentially leading to inflated confidence in its out-of-sample prediction performance. On the other hand, although the physics-informed neural networks (PINNs) (*23-27*) have been increasingly used to improve physically relevant results, few deep learning methods have been reported to embed fluid physics-constraints to improve air quality assessment. Hahnel et al. (*28*) used deep leaning to extend the range of air pollution monitoring and forecast.

We found that encoding fluid dynamics physics in deep learning frameworks substantially reduces estimation bias in *jointly* predicting $NO_2$ and $NO_x$ (Materials and Methods). Our modeling framework is based on a 2-dimensional (2-D) terrain-following Eulerian system (Fig. 2A). The principles of atmospheric fluid dynamics incorporated in this model included the conservation of mass and continuity principles for the atmosphere (*29*). The former assumes that the total mass of an air pollutant within a control volume is conserved in the balance given the inflows, outflows, emission sources and sinks during a period, and the latter assumes that the atmospheric fluid is continuous across space (i.e., it contains no holes). For a grid cell or target location, the continuity equation simulates the temporal evolution of $NO_2$ or $NO_x$ concentrations, taking into account the pollutant sources, transport, sinks and meteorological, topological and geographical effects (*30*). Based on automatic differentiation learning (*31*), the physics-informed neural network (PINN) method we developed can address multiple spatial scales (*32*). Using Reynolds decomposition (*33*), advection and diffusion processes are summarized based on velocities, partial derivatives for advection, diffusion coefficients, and eddy diffusion terms. To simulate the fluid fields, partial differential equations (PDEs) are included in the PINN model to explicitly characterize 2-D advection and diffusion (*23*) (Fig. 2C-F). Due to the lack of vertically distributed measurements of $NO_x$ concentrations (i.e. all available measurements were at ground-level), we could not consider vertical $NO_x$ advection, but we did include vertical meteorological reanalysis data such as air temperature and wind speed at different altitudes (2m, 10m, 50m), total $NO_2$ columns, and planetary boundary layer height (PBLH) to account for the influence of vertical processes on ground-level $NO_2$ and $NO_x$ concentrations. Thus, we primarily simulated horizontal advection and diffusion along orthogonal directions (latitude and longitude) (Fig. 2D). In addition, given the significant impact of complex terrain on local dispersion of ground-level air pollutants (Fig. 1F) (*34*), we also simulated the vertical dispersion of $NO_x$ within the range of ground elevations of our samples (Fig. 2E). While the continuity equation was used to simulate fine-scale transport of $NO_x$, we also introduced another compound parameter in the model to account for the balanced changes in emissions, chemical transformations, and sinks.

In our models, we borrowed concepts from CTMs and simplified the parameterizations for wind velocity, diffusion coefficients and other parameters using automatic differentiation learning. The $NO_x$ concentration data alone do not provide sufficient information to generate a single solution for these unknown parameters (i.e., the Reynolds number), but it is impossible to acquire such parameter data in practice and solving inverse problems with hidden physics is prohibitively difficult and expensive (*23*). As shown in the model with hidden fluid mechanics (*26*), we also achieved a high quantitative agreement between observed and predicted proxy values of these physical parameters by relying solely on the information contained in the observed concentration data. Here, we constructed a physics-informed neural network to approximate the following mapping (Fig. 2): $(t,x,y,z,E,T,M,L,O) \rightarrow (C,C';\theta,\theta')$ (*t*: time, *x*: longitude, *y*: latitude, *z*: elevation, *E*: emission or its proxy, *T*: transport, *M*: meteorology, *L*: land-use; *O*: other influential factors; $C$ ($C'$): $NO_2$ ($NO_x$) concentration, $\theta$ ($\theta'$): parameters (velocity, diffusion coefficient and Reynolds number etc.$_2$ ($NO_x$)). Two full residual deep networks (Fig. 2B-C, fig. S1) were constructed for parameter estimation (Fig. 2C) in support of PDE residuals, and joint predictions (Fig. 2B) of $NO_2$ and $NO_x$, respectively. These physical parameters have been reported to be affected by emission, meteorology, topography, geography and land-use (*30*). Full residual deep networks have also been used to improve efficiency in regression learning (*35*). The PDEs of $NO_2$ and $NO_x$ were encoded using two residuals, $e_1$ and $e_2$ (Fig. 2-F). The PDE residuals encouraged or guided the model in training to preserve the continuity equation as much as possible. PINN encouraged all data (including training and predicting samples) to conform to the mass conservation in support of the PDE residuals. Furthermore, residual multilayer perceptron (MLP) was used as a general nonlinear approximator for any continuous function (*24, 36*) for computational efficiency and since it could simulate the continuity of air well.

As described above, $NO_2$ has complex atmospheric physiochemical relationships with NO and other air pollutants (Fig. 2I). As a component of $NO_x$, the $NO_2$ concentration is always less than or equal to the $NO_x$ concentration, so to represent this in a unified model, the joint output of $NO_2$ and $NO_x$ was set up so that the input and the weight parameters of the constructed neural networks could be shared between $NO_2$ and $NO_x$ to represent their interaction efficiently. Similar sharing has been used to improve multitask machine learning (*37, 38*). In addition, the thresholds (maximums based on actual observed data and theoretical concentration limits) for $NO_2$ and $NO_x$ concentrations were encoded via the residuals, $e_3$ and $e_4$, and their concentration relationship ($C(NO_2) \leq C(NO_x)$) was encoded via the residual, $e_5$. All data satisfied the thresholds and relationships ($e_1$-$e_5$). In addition, learning using training samples minimized mean squared error (MSE) residuals between observed and predicted (using $e_6$ and $e_7$, Fig. 2H) $NO_2$ and $NO_x$ values.

To maintain the loss function as sufficiently differentiable, the total loss function ($\mathcal{L}$) is defined as the sum of the mean squared residuals multiplied by their respective weights ($\lambda_i$) (*39*):

$$\mathcal{L} = \frac{1}{N}\sum_{i=1}^{5}\sum_{j=1}^{N}\lambda_i e_i(\mathbf{X}_j)^2 + \frac{1}{M}\sum_{i=6}^{7}\sum_{j=1}^{M}\lambda_i e_i(\mathbf{X}_j)^2 \quad (1)$$

where $N$ denotes the number of all samples except the site-based independent samples, $M$ denotes the number of training samples, and $X_j$ is the input matrix (see Eq. S1-S9 in Materials and Methods for details).

We used the nonparametric bootstrap method to assess the uncertainty of the predictions since it performs as well as parametric uncertainty and works for non-normal and nonlinear data (*40*). Usually, generalization error is defined as the error of the model on all the data (including data at unseen locations) within the spatiotemporal domain, and consists of estimation bias, variance and random noise (*38*). We used the training and testing samples to derive the bias, and all the data (training and testing samples, as well as predictions) to derive the variance, and then derive an estimate of the generalization error according to the following equation (Materials and Methods):

$$\varepsilon_G = \int_{\mathcal{T}}\int_{\mathcal{S}}\|\mathbf{u}(\mathcal{S},t)-\mathbf{u}^*(\mathcal{S},t)\|d\mathcal{S}dt = \eta_{\mathbb{D}}^b + \eta_{\mathbb{D}}^v + \varepsilon_{\mathbb{D}}^n \approx \eta_{\mathbb{S}}^b + \eta_{\mathbb{S}}^v + \varepsilon_{\mathbb{S}}^n \quad (2)$$

where $\mathbb{D} = L^3(\mathcal{S}\times\mathcal{T})$ ($\mathcal{S} = \mathcal{X}\times\mathcal{Y}\times\mathcal{Z}$ denotes 3-D spatial domain of ground surface [$\mathcal{X}$: longitude, $\mathcal{Y}$: latitude, $\mathcal{Z}$: elevation], $\mathcal{T}$ denotes temporal domain), $\int_{\mathcal{S}}\cdots$ is a shorthand of $\int_{\mathcal{X}}\int_{\mathcal{Y}}\int_{\mathcal{Z}}\cdots$, and $\mathbf{u}^*(\mathcal{S},t)$ denotes the trained PINN, approximating the observed values, $\mathbf{u}(\mathcal{S},t)$;

$\eta_\mathbb{D}^b$ denotes the model/estimation bias, $\eta_\mathbb{D}^v$ denotes the model variance, and $\varepsilon_\mathbb{D}^n$ denotes random noise, $\varepsilon_\mathbb{D}^n \sim \mathcal{N}(0, \sigma^2)$, $\eta_\mathbb{S}^b$, $\eta_\mathbb{S}^v$ and $\varepsilon_\mathbb{S}^n$ denotes the model bias, variance and random noise in the sample space, respectively.

Employing PDEs ($e_1$ and $e_2$) and additional physical constraints ($e_3$-$e_5$) in the residuals allowed us to encode physics in deep learning. The mini-batch gradient descent optimizer Adam enabled the trained model to penalize the residual formula for nearly "infinite" samples to conform to the laws of fluid physics (*26*). In practice, instead of the regular grid system simulated by CTM models, our variable-spacing method has greater flexibility for simulation of the fine-and coarse-scale transport of $NO_2$ and $NO_x$. Furthermore, based on the principles of PDE conditional stability and quadrature rules, we observed in our experimental results that utilizing a substantial number of unsupervised and/or supervised samples to regulate the model parameters in accordance with the physical laws can lead to a reduction in generalization error. This reduction brings the generalization error closer to the training/testing error (*41*). We also validated the trained model by site-based independent testing that used data from locations that were not used in model training (Materials and Methods). Further, using the joint physics-informed neural network (jPINN) as the base model, the meta-algorithm of non-parametric bootstrap aggregation (bagging) (*42*) was used to generate ensemble predictions and uncertainty estimates (*43*). We derived model and sampling bias from training and testing errors and obtained model variance using bagging predictions (fig. S2A). Specifically, the model bias was estimated for different levels of air pollutant concentrations from the training, regular testing and site-based testing errors (fig. S2B). The findings demonstrated that our model's generalization error closely aligned with both the training and testing errors, as well as the site-based testing error. This alignment was in line with theoretical proof [Eq. 2], and further supports the proximity of our proposed uncertainty estimate to the actual generalization error.

We selected California as the testbed for demonstrating the method (fig S2). In total, 81,929 $NO_2$ and $NO_x$ samples of weekly average measurements from 2004 to 2020 (17 years) and 1,342 sampling locations were used. These data were from ~150 long-term government monitoring stations (*19*) and 1,192 locations used in short-term saturation sampling campaigns (*44-47*). Most of the saturation sampling locations were highly clustered and captured fine-scale spatial characteristics. About 81% of all sampling locations were selected for training and regular testing, and the remaining 19% were used as site-based independent testing samples. For the 81% sampling locations, the data were sampled in a mixture of spatial and temporal dimensions with county and season as a combined stratifying factor, and 78% of these samples were used for training and the remaining 22% were used for regular testing. We chose the sample so that the proportion of the total samples for regular and site-based tests was similar to the proportion (~0.368) of the unique test samples in the bootstrap 0.632 rule (*48*). To evaluate the generalization of the trained models, 150 base models were trained using the same bootstrap samples, and the statistics and violin plots of R-squared ($R^2$) and Root Mean Square Error (RMSE) were compared between six representative methods, including jPINN (our model), jPINN with no elevation used in the PDE (our model removing z term in the PDE), separate PINN for single prediction of $NO_2$ or $NO_x$, XGBoost, random forest and full residual neural network (FRNN (*35*)). The latter three machine learning methods were chosen since they are increasingly used for air quality prediction due to their high accuracy and computational efficiency. For all the models, proxies to emissions data were used to represent traffic ($NO_x$ output by CALINE4 dispersion model (*49*), traffic density, road intersection (https://www.here.com), distance to major roadways), restaurant locations from OpenStreetMap, and satellite data (NDVI). We used meteorological data from the gridMET of contiguous US and MERRA2 reanalysis. We also introduced the NASA's MINDS reanalysis variables (*50*) as additional input to account for background pollutant transformation. Furthermore, we also included aircraft emissions, land-use data and satellite bands to account for emission and/or deposition factors. See Materials and Methods for details of all data and modeling details, including other methods compared (table S1).

Performance statistics of 150 base models show that jPINNs achieved the best generalization, as shown by high mean $R^2$ (0.95-0.96) and small RMSE (1.54 ppb for $NO_2$; 3.25 ppb for $NO_x$) (table S2) from site-based independent tests, as well as small variance in these statistics (Fig. 3AB and fig. S6). Although XGBoost, random forest and full residual neural network reported similar training ($R^2$: 0.92-0.98) and regular testing ($R^2$: 0.88-0.92) metrics, they had low or moderate site-based independent test results ($R^2$: 0.53-0.75; RMSE: 2.96-14.74 ppb for $NO_2$ and 7.06-41.42 ppb for $NO_x$). The large difference (20-43% in mean $R^2$ and 2.27-3.97 ppb [$NO_2$] / 7.06-11.10 ppb [$NO_x$] in mean RMSE) in independent test statistics illustrates the improved performance of the jPINNs compared to purely data-driven ML models and suggest their true generalizability might be overly optimistic and severely overestimated. In comparison, the jPINNs achieved the lowest generalization error (RMSE: 1.54 ppb for $NO_2$ and 3.25 ppb for $NO_x$) with small difference between training/regular testing and site-based testing, substantially reducing estimation bias. Furthermore, the comparisons with separate (single ensemble run) PINNs show that the joint training of $NO_2$ and $NO_x$ distributions in jPINN better capture their complex physicochemical interactions, resulting in more stable learning (smaller variance, Fig. 3) and significantly improved generalizability (on average by 8-16% in $R^2$). Also, the elevation PDE improved generalization, increasing $R^2$ by about 1-3% and decreasing RMSE by about 0.13 to 0.98 ppb, compared to not including it. Our results also show (Fig. 3C) that the physical loss terms ($e_1$-$e_5$) progressively reached a minute value close to zero as the learning process proceeded. This outcome indicated that our models were encouraged to preserve mass conservation ($e_1$: $1.49 \times 10^{-5}$; $e_2$: $1.52 \times 10^{-5}$), comply with the threshold constraints for $NO_2$ and $NO_x$ ($e_3$:0; $e_4$:0), as well as maintain their physical relationships ($e_5$: $2.39 \times 10^{-3}$).

Scatterplots of the residuals versus predicted values further showed the jPINNs had superior performance (Fig. 3DE) compared to bagging baseline FRNN. Our jPINN substantially improved generalization, increasing $R^2$ by 26-33% and decreasing RMSE by 2.89 ppb ($NO_2$) and 7.87 ppb ($NO_x$) in site-based independent testing. On average, the model variance was about 11-13% of the generalization error, with the remaining 87-89% including bias and random noise (fig. S7). Therefore, bias is the main source of generalization error since random noise is assumed to have a normal distribution with zero means. A comparison of $NO_2$ to $NO_x$ concentration ratios (fig. S8) showed that all the estimates by jPINNs satisfied the physical relationship ($C \leq C'$) between $NO_2$ and $NO_x$, while bagging predicted concentrations of FRNNs did not. The time series of weekly tests (fig. S9-S10) revealed that although the baseline FRNNs demonstrated strong performance in regular tests, they consistently underperformed in each weekly site-based test. This can be attributed to the absence of physical laws in the models. The FRNNs consistently overestimated or underestimated observations at a location under the study domain (fig. S11 for an example of site overestimation). The other purely data-driven machine learning methods have similar systematic bias. The reliance of ML models on reproducing existing relationships in the data does not seem to enable the trained models to adequately capture the spatiotemporal evolution of $NO_x$, potentially leading to deviations from the physical laws of mass conservation and continuity, and introducing systematic bias. Under such systematic bias, exposure misclassification can occur, potentially resulting in misleading findings for downstream applications such as in air pollution epidemiologic studies (*51-53*). Sensitivity analysis also showed that the jPINNs were robust to missingness in coarse-scale MINDS pollutants and land-use covariates (24 covariates in total; just a slight change in RMSE (≤0.5 ppb) and in $R^2$ (≤1%); Fig. 4H-I, fig. S12-S13).

Tree-based ML methods (e.g., random forest (*54*), XGBoost (*55*), and extreme random trees (*56*)) use binary discretization which may result in discontinuity of predictions, thus violating the continuity law of fluid dynamics and possibly leading to discontinuity bias (*35*) compared to our approach. Our jPINN prediction maps (Fig. 4) captured expected spatial patterns from meteorological fields and important local sources of traffic and roads as seen by local hotspots of $NO_2$ concentrations in low altitude terrains, high traffic intersections, and during low PBLH, and weak atmospheric turbulence. Under similar meteorology (slightly higher wind speed for 2019, fig. S14), the evolution of concentrations predicted by the jPINN from February 19 to March 25, 2020 (fig. S15, S16) accurately captured the continued decline in California $NO_2$ emissions compared to 2019 (fig. S17, S18) probably due to implementation of shelter-in-place orders against the coronavirus. In contrast to previous studies that did not provide accurate uncertainty estimation, our approach successfully derived uncertainty estimates (Fig. 4C-D) with narrow 95% confidence intervals. Additionally, we employed a statistically innovative method to achieve a high coverage probability (0.95) for the prediction interval of $NO_2$ and $NO_x$. This demonstrates that our uncertainty estimates accurately captured the level of uncertainty in the predictions.

As shown in the feature importance analysis in our jPINNs (fig. S19), the emission sources or their proxy variables (BOT $NO_2$, $O_3$, traffic and landuse related variables) made important contributions to their prediction (≥20% of explained variance); as the driving and influential factors for all air pollutants, meteorology accounted for ≥30% of explained variance. By maintaining the governing physical laws of mass conservation for all the samples (including prediction points and those beyond the observation dataset), the trained model presented robust prediction accuracy (low estimation bias), less affected by the values outside the training samples in predictions. In conclusion, our jPINN method was able to achieve greater prediction generalization, compared to well-established other data driven ML models and is trained in an efficient end-to-end manner that could potentially be more scalable compared to running CTMs. Thus, it provides flexibility and applicability for air pollution estimation problems where many of the control variables are unmeasured. Comparing with early PINNs (*23*) and forecast (*28*), we considered the complexity and diversity of influencing factors, encoded shared physical laws, and extracted critical coefficients (e.g., velocity, diffusion) from the covariates through efficient learning. By jointly modeling $NO_2$ and $NO_x$, encoding physics knowledge directly into deep learning, and quantifying uncertainty, we have successfully attained prediction performance with significantly reduced bias and increased interpretability and generalizability in our models. In doing so, we have exemplified how the incorporation of physiochemical knowledge can serve as a key advancement for future improvements in machine learning models for air pollution exposure estimation. This study offers significant improvements to existing machine learning methods for unprecedented reduction in the bias for reliable estimation of $NO_x$, which can be also applied for other air pollutants or similar atmospheric environmental predictions.

**Acknowledgments:** The authors gratefully thank Dr. Melanie Follette-Cook, Dr. Pawan Gupta, Dr. Bryan Duncan, Ms. Abigail Nastan, Dr. Sina Hasheminassab, and Dr. David Diner for guidance and insights on NASA data products and their application in health and air quality studies. Finally, the authors gratefully acknowledge the support of NVIDIA Corporation with the donation of the Titan Xp GPUs used in this research.

**Funding:** The study was supported by the Lifecourse Approach to Developmental Repercussions of Environmental Agents on Metabolic and Respiratory Health NIH ECHO grants (4UH3OD023287) and the Southern California Environmental Health Sciences Center (National Institute of Environmental Health Sciences' grant, P30ES007048).

**Author contributions:** L.L. and R.H. designed the project; L.L. conceived the original idea and conducted critical method development and data analyses; F.L. and N.P. performed the extraction of the covariate data including transportation, land-use and emission etc.; R.H., J.W. and B.R. provided the data support of field measurements; L.O. conducted and extracted the MINDS simulation and provided its data for use in this project; R.H. supervised the overall project implementation. All authors contributed to developing the theory, providing feedback on the analysis, and writing the manuscript.

**Competing interests:** Authors declare that they have no competing interests.

**Data and materials availability:** Model code will be publicly available following publication. The original field measurement samples cannot be shared due to the data use protocol. The ambient air monitoring data used in training is publicly available from the EPA.

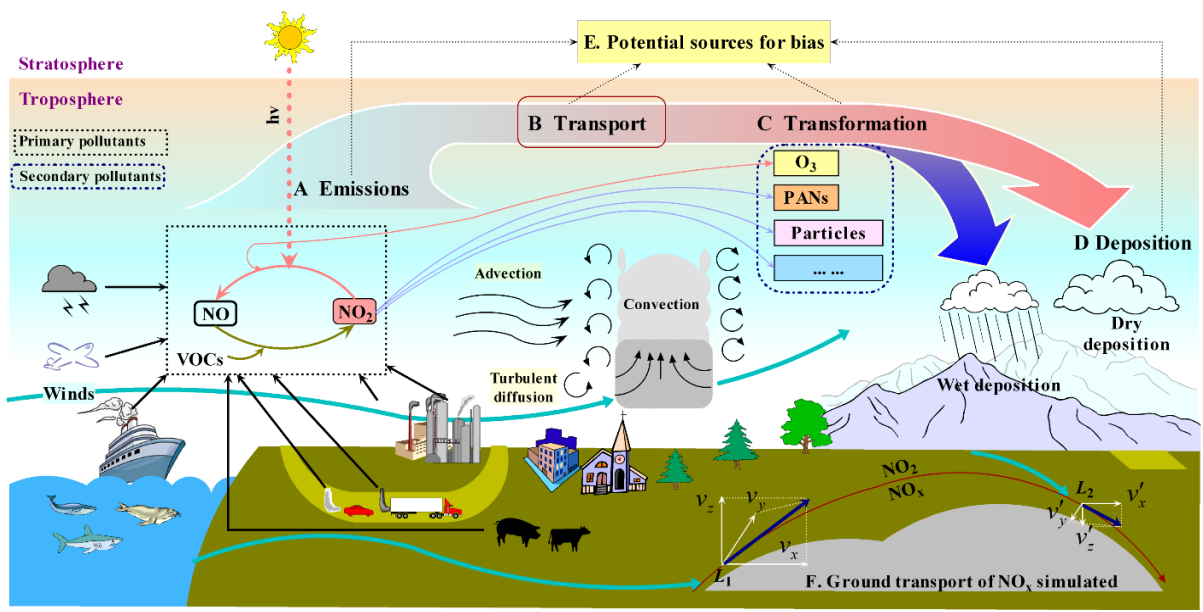

**Fig. 1. The emission, transport, transformation and deposition processes of NO$_x$ (NO$_2$ and NO).** (**A**) Emissions of NO$_x$ include primarily anthropogenic sources (cars, trucks, ships, power plants, off-road equipment, etc.), as well as secondary natural sources (wildfires, biomass burning and lightning etc.). The majority of NO$_2$ is formed in the air by VOCs and NO. With sunlight, NO$_2$ may convert back to NO. (**B**) Transport of NO$_2$ in the advection and diffusion way due to the influence of prevailing wind. (**C**) Atmospheric chemical transformation of NO and NO$_2$ due to the meteorology (e.g., sunlight) and other transported pollutants. (**D**) Deposition of NO$_2$ and NO$_x$ by rain, snow, fog, or particles etc. (**E**) Potential bias sources for NO$_2$ and NO$_x$ prediction from emission, transport, transformation and deposition. (**F**) Ground transport of NO$_x$ simulated by the model.

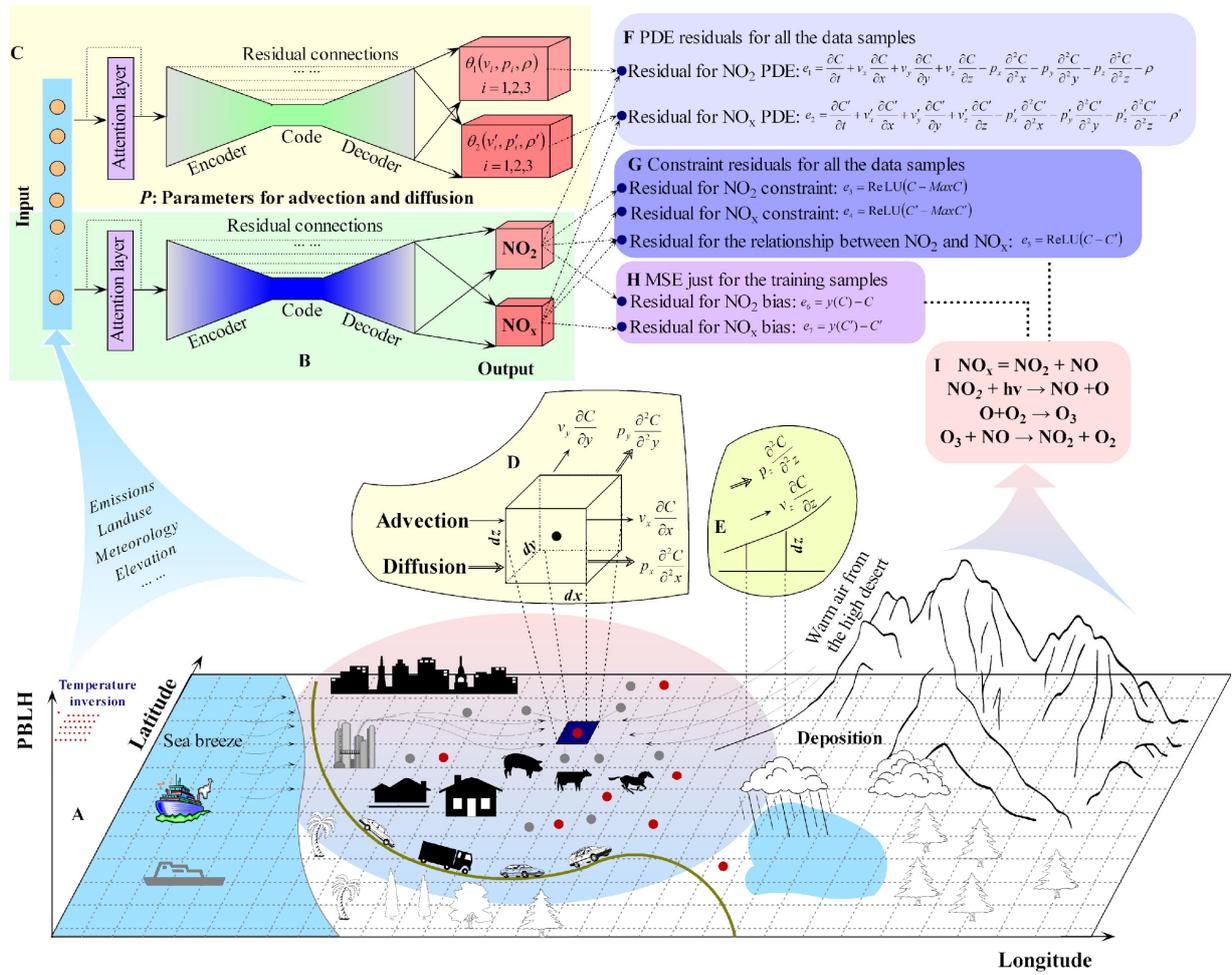

**Fig. 2. The semi-supervised modeling framework of residual deep learning by encoding the advection and diffusion of $NO_2$ and $NO_x$ and their relationships.** (**A**) The Eulerian system of regular 2-D grids for simulating the fine-scale advection and diffusion of $NO_2$ and $NO_x$. The air temperature inversion may trap $NO_x$ under a lower mixed layer and aggregate air pollution on the ground. (**B** and **C**) The joint deep learning system consisting of two residual networks respectively for estimating the parameters (C) for the transport of $NO_2$ and $NO_x$, and estimating $NO_2$ and $NO_x$ concentrations (B). (**D**) Simulation of the horizontal advection and diffusion in the atmospheric continuity equation. (**E**) Simulation of the advection and diffusion to account for variation of $NO_2$ and $NO_x$ by surface elevation. (**F**) Residuals for the advection and diffusion of $NO_2$ ($e_1$) and $NO_x$ ($e_2$) using PDE. (**G**) Residuals for the maximum constraints of $NO_2$ ($e_3$) and $NO_x$ ($e_4$), and their physical relationship ($e_5$). (**H**) MSE residuals for $NO_2$ ($e_6$) and $NO_x$ ($e_7$) bias. (**I**) The complex physiochemical reactions between $NO_2$ and $NO_x$ captured by the joint system.

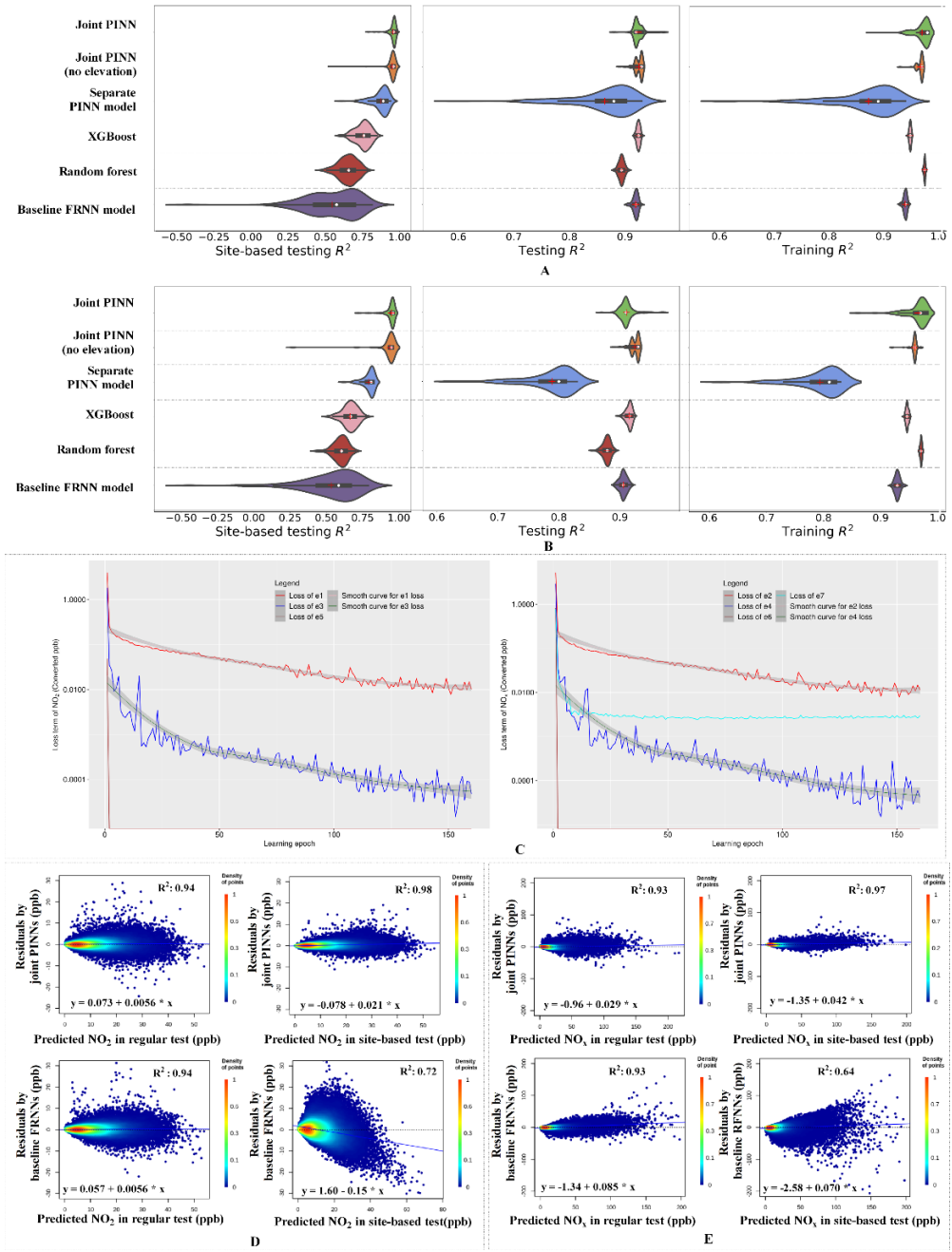

**Fig. 3. Comparison of generalization of $NO_2$ and $NO_x$ between physics-informed deep learning and representative machine learning methods.** (**A** and **B**) Violins of $R^2$ in site-based testing, regular testing and training for $NO_2$ (A) and $NO_x$ (B). (**C**) Learning curves for loss terms. (**D** and **E**) Comparison of the scatter plots of residuals vs. ensemble predicted values between pure deep learning and physics-informed deep learning for $NO_2$ (D) and $NO_x$ (E), and the variance explained improved by 26% ($NO_2$) and 33% ($NO_x$) in site-based independent testing. PINN: physics-informed neural network. FRNN: full residual neural network. Red plus indicates mean $R^2$.

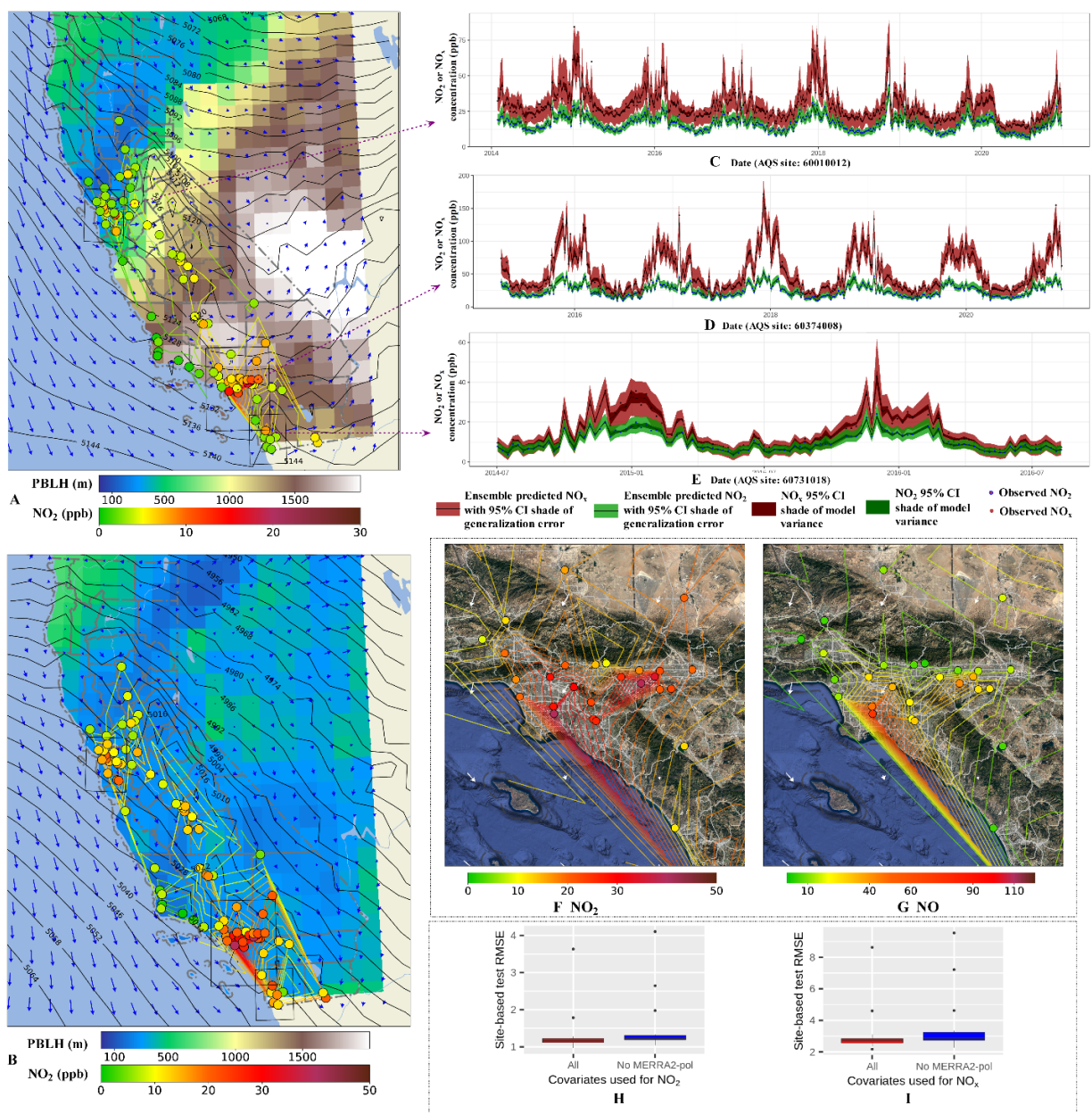

**Fig. 4. Ensemble predicted NO$_2$ and NO$_x$, their gradients, uncertainty analysis.** (**A** and **B**) The ensemble predicted NO$_2$ with PBLH (black isolines) and wind vector (blue arrows) for a summer (A, June 24-30, $R^2$: 0.98; RMSE: 0.51 ppb) and winter (B, Dec. 16-22, $R^2$: 0.97; RMSE: 1.35 ppb) week of 2020. (**C-E**) Time series of the ensemble predicted NO$_2$ and NO$_x$ with their 95% confidence intervals of generalization error and model variance for three AQS routine monitoring locations in San Francisco (C), Los Angeles (D) and San Diego (E). (**F-G**) Ensemble predicted NO$_2$ and NO in a winter week of 2020 in the Los Angeles area. (**H-I**) Comparison of site-based test RMSE of the full model with that without the MINDS pollutant input. The true color high-res satellite image is used as the background for F-K. NO was defined as the difference between the predicted NO$_x$ and NO$_2$ concentrations.

# References

1. T. W. Hesterberg *et al.*, Critical review of the human data on short-term nitrogen dioxide (NO2) exposures: Evidence for NO2 no-effect levels. *Crit Rev Toxicol* **39**, 743-781 (2009).
2. I. C. Mills, R. W. Atkinson, S. Kang, Quantitative systematic review of the associations between short-term exposure to nitrogen dioxide and mortality and hospital admissions (vol 5, e006946, 2015). *Bmj Open* **5**, (2015).
3. G. Weinmayr, E. Romeo, M. De Sario, S. K. Weiland, F. Forastiere, Short-Term Effects of PM10 and NO2 on Respiratory Health among Children with Asthma or Asthma-like Symptoms: A Systematic Review and Meta-Analysis. *Environ Health Persp* **118**, 449-457 (2010).
4. A. Faustini, R. Rapp, F. Forastiere, Nitrogen dioxide and mortality: review and meta-analysis of long-term studies. *Eur Respir J* **44**, 744-753 (2014).
5. D. M. Stieb *et al.*, Systematic review and meta-analysis of cohort studies of long term outdoor nitrogen dioxide exposure and mortality. *PloS one* **16**, (2021).
6. P. M. Edwards *et al.*, Transition from high- to low-NOx control of night-time oxidation in the southeastern US. *Nat Geosci* **10**, 490-+ (2017).
7. EPA. (2022).
8. B. Zhao *et al.*, NOx emissions in China: historical trends and future perspectives. *Atmos Chem Phys* **13**, 9869-9897 (2013).
9. Y. B. Zhao *et al.*, Substantial Changes in Nitrogen Dioxide and Ozone after Excluding Meteorological Impacts during the COVID-19 Outbreak in Mainland China. *Environ Sci Tech Let* **7**, 402-408 (2020).
10. Y. Zhu *et al.*, Impacts of TROPOMI-Derived NO X Emissions on NO2 and O3 Simulations in the NCP during COVID-19. *ACS Environmental Au*, (2022).
11. B. J. Finlayson-Pitts, J. N. Pitts Jr, *Chemistry of the upper and lower atmosphere: theory, experiments, and applications*. (Elsevier, 1999).
12. X. Lu, L. Zhang, L. Shen, Meteorology and Climate Influences on Tropospheric Ozone: a Review of Natural Sources, Chemistry, and Transport Patterns. *Curr Pollut Rep* **5**, 238-260 (2019).
13. WHO, "Review of evidence on health aspects of air pollution—REVIHAAP project: Final technical report," (The WHO European Centre for Environment and Health, Bonn, Switzerland, 2013).
14. WHO. (2021).
15. J. Richmond-Bryant *et al.*, Estimation of on-road NO2 concentrations, NO2/NOX ratios, and related roadway gradients from near-road monitoring data. *Air Qual Atmos Hlth* **10**, 611-625 (2017).
16. A. A. Karner, D. S. Eisinger, D. A. Niemeier, Near-roadway air quality: synthesizing the findings from real-world data. *Environmental science & technology* **44**, 5334-5344 (2010).
17. X. D. Zou *et al.*, Shifted power-law relationship between NO2 concentration and the distance from a highway: A new dispersion model based on the wind profile model. *Atmospheric Environment* **40**, 8068-8073 (2006).
18. M. Kozdron. (2016).
19. L. Li *et al.*, Constrained Mixed-Effect Models with Ensemble Learning for Prediction of Nitrogen Oxides Concentrations at High Spatiotemporal Resolution. *Environmental Science and Technology* **(in press)**, (2017).
20. A. Daly, P. Zannetti, in *Ambient Air Pollution,* D. Al-Ajmi, A. Al-Rashied, Eds. (The Arab School for Science and Technology (ASST) and The EnviroComp Institute, 2007).
21. C. Bellinger, M. S. Mohomed Jabbar, O. Zaiane, A. Osornio-Vargas, A systematic review of data mining and machine learning for air pollution epidemiology. *BMC Public Health* **17**, 907 (2017).
22. M. M. Kelp, D. J. Jacob, J. N. Kutz, J. D. Marshall, C. W. Tessum, Toward Stable, General Machine-Learned Models of the Atmospheric Chemical System. *J Geophys Res-Atmos* **125**, (2020).
23. G. E. Karniadakis *et al.*, Physics-informed machine learning. *Nat Rev Phys* **3**, 422-440 (2021).
24. L. Lu, P. Jin, G. Pang, Z. Zhang, G. E. Karniadakis, Learning nonlinear operators via DeepONet based on the universal approximation theorem of operators. *Nature machine intelligence* **3**, 218-229 (2021).
25. M. Raissi, P. Perdikaris, G. E. Karniadakis, Physics-informed neural networks: A deep learning framework for solving forward and inverse problems involving nonlinear partial differential equations. *J Comput Phys* **378**, 686-707 (2019).
26. M. Raissi, A. Yazdani, G. E. Karniadakis, Hidden fluid mechanics: Learning velocity and pressure fields from flow visualizations. *Science* **367**, 1026-+ (2020).
27. S. Wang, H. Wang, P. Perdikaris, Learning the solution operator of parametric partial differential equations with physics-informed DeepONets. *Science advances* **7**, eabi8605 (2021).

28. P. Hahnel, J. Marecek, J. Monteil, F. O'Donncha, Using deep learning to extend the range of air pollution monitoring and forecasting. *J Comput Phys* **408**, (2020).
29. D. Andrews, *An Introduction to Atmospheric Physics*. (Cambridge University, Cambridge 2010).
30. Z. M. Jacobson, *Fundamentals of Atmospheric Modeling, 2nd Edition*. (Cambridge University Press, 2005).
31. G. A. Baydin, B. Pearlmutter, A. A. Radul, J. Siskind, Automatic differentiation in machine learning: a survey. *Journal of Machine Learning Research* **18**, 1-43 (2018).
32. EPA. (2016).
33. J. Pedlosky, *Geophysical fluid dynamics*. (Springer, 1987).
34. D. C. Whiteman, in *Mountain Meteorology: Fundamentals and Applications,* D. C. Whiteman, Ed. (, Oxford University Press 2000).
35. L. Li, Y. Fang, J. Wu, J. Wang, G. Y., Encoder-Decoder Full Residual Deep Networks for Robust Regression Prediction and Spatiotemporal Estimation. *IEEE Transactions on Neural Networks and Learning Systems* **32**, 4217-4230 (2021).
36. K. Hornik, Approximation capabilities of multilayer feedforward networks. *Neural Networks* **4**, 251-257 (1991).
37. Z. Zhang, P. Luo, C. C. Loy, X. Tang, Learning Deep Representation for Face Alignment with Auxiliary Attributes. *IEEE Trans Pattern Anal Mach Intell* **38**, 918-930 (2016).
38. M. C. Bishop, *Pattern Recognition and Machine Learning*. (Springer, 2006).
39. K. Mahendru. (2019).
40. S. Kumar, A. Srivastava, in *Proc. 18th ACM SIGKDD Conf. Knowl. Discovery Data Mining*. (Association for Computing Machinery, Beijing, 2012), pp. 08/12/2012-2008/2016/2012.
41. S. Mishra, R. Molinaro, Estimates on the generalization error of physics-informed neural networks for approximating a class of inverse problems for PDEs. *IMA Journal of Numerical Analysis* **42**, 981-1022 (2022).
42. L. Breiman, Bagging Predictors. *Machine Learning* **24**, 123-140 (1996).
43. S. Kumar, A. N. Srivistava, in The 18th ACM SIGKDD Conference on Knowledge Discovery and Data Mining. (2012).
44. M. Franklin *et al.*, Predictors of intra-community variation in air quality. *J Expo Sci Environ Epidemiol* **22**, 135-147 (2012).
45. S. Fruin *et al.*, Spatial Variation in Particulate Matter Components over a Large Urban Area. *Atmos Environ (1994)* **83**, 211-219 (2014).
46. L. Li, J. Wu, M. Wilhelm, B. Ritz, Use of generalized additive models and cokriging of spatial residuals to improve land-use regression estimates of nitrogen oxides in Southern California. *Atmos Environ (1994)* **55**, 220-228 (2012).
47. J. G. Su *et al.*, Predicting traffic-related air pollution in Los Angeles using a distance decay regression selection strategy. *Environ Res* **109**, 657-670 (2009).
48. D. Berrar, W. Dubitzky, in *Encyclopedia of Systems Biology,* W. Dubitzky, O. Wolkenhauer, K. Cho, H. Yokota, Eds. (Springer, New York, NY, 2013).
49. P. Benson, K. BPinkerman, "CALINE4,-A Dispersion Model for Predicting Air Pollutant concentrations near Roadways. ,"  (1989).
50. L. N. Lamsal *et al.*, Ozone Monitoring Instrument (OMI) Aura nitrogen dioxide standard product version 4.0 with improved surface and cloud treatments. *Atmos Meas Tech* **14**, 455-479 (2021).
51. L. Sheppard *et al.*, Confounding and exposure measurement error in air pollution epidemiology. *Air Qual Atmos Health* **5**, 203-216 (2012).
52. X. Wu *et al.*, Causal Inference in the Context of an Error Prone Exposure: Air Pollution and Mortality. *Ann Appl Stat* **13**, 520-547 (2019).
53. S. L. Zeger *et al.*, Exposure measurement error in time-series studies of air pollution: concepts and consequences. *Environ Health Perspect* **108**, 419-426 (2000).
54. X. Hu *et al.*, Estimating PM2.5 Concentrations in the Conterminous United States Using the Random Forest Approach. *Environmental science & technology* **51**, 6936-6944 (2017).
55. P.-Y. Wong *et al.*, Using land-use machine learning models to estimate daily NO2 concentration variations in Taiwan. *Journal of Cleaner Production* **317**, 128411 (2021).
56. J. Wei *et al.*, Ground-level NO2 surveillance from space across China for high resolution using interpretable spatiotemporally weighted artificial intelligence. *Environmental science & technology* **56**, 9988-9998 (2022).

# Supplementary Materials for

## Physics-Informed Deep Learning to Reduce the Bias in Joint Prediction of Nitrogen Oxides

Lianfa Li[1,2*], Roxana Khalili[1], Frederick Lurmann[3], Nathan Pavlovic[3], Jun Wu[4], Yan Xu[1], Yisi Liu[1], Karl O'Sharkey[1], Beate Ritz[5], Luke Oman[6], Meredith Franklin[1,7], Theresa Bastain[1], Shohreh F. Farzan[1], Carrie Breton[1], Rima Habre[1,8*]

[1]Department of Population and Public Health Sciences, University of Southern California, Los Angeles, CA, USA.
[2]State Key Laboratory of Resources and Environmental Information System, Institute of Geographical Sciences and Natural Resources, Chinese Academy of Sciences, Beijing, China.
[3]Sonoma Technology, Inc., Petaluma, CA, USA.
[4]Program in Public Health, Susan and Henry Samueli College of Health Sciences, University of California, Irvine, CA, USA.
[5]Departments of Epidemiology and Environmental Health, Fielding School of Public Health, University of California, Los Angeles, CA, USA.
[6]Goddard Space Flight Center, National Aeronautics and Space Administration, Greenbelt, MD, USA.
[7] Department of Statistical Sciences, University of Toronto, Toronto, Ontario Canada.
[8] Spatial Sciences Institute, University of Southern California, Los Angeles, CA, USA.

*Correspondence to: lianfali@usc.edu (L.L.); habre@usc.edu (R.H.).

**This PDF file includes:**

    Materials and Methods
    Supplementary Text
    Figs. S1 to S19
    Tables S1 to S2

**Materials and Methods**

<u>Modeling architecture</u>

The flexible modeling architecture can be applied for the Eulerian system of regular 2-D grids at a fine spatial (e.g., 50 m for our case) and temporal (e.g., weekly for our case) scale (Fig. 2A) and a meshfree system of point samples. According to the assumptions of conservation of mass and continuity underlying atmospheric motions, the change of $NO_2$ or $NO_x$ concentration for a grid cell or target location can be simulated using the following continuity equation:

$$\frac{\partial C}{\partial t} = -\nabla(VC) + p\nabla^2 C + R \tag{S1}$$

where $C$ is $NO_2$ or $NO_x$ concentration to be predicted, $\partial t$ denotes the time derivative, $\partial C / \partial t$ denotes $NO_2$ or $NO_x$ change in each grid cell or target location over time ($t$), $\nabla(VC)$ denotes advection, $V$ denotes velocity, $p$ denotes the coefficient for the diffusion term ($\nabla^2 C$), $R$ denotes the total change by chemical transformation, emission and deposition.

Given the lack of vertical measurements of concentrations (i.e. measurements at different heights above ground-level), the convection simulation along the atmospheric vertical profile was not conducted. However, parameters on vertical winds and planetary boundary layer height from reanalysis and meteorological data were included within the model to account for the influence of vertical mixing and processes on ground $NO_x$ concentrations. Given the significant influence of the terrain on the dispersion of air pollutants, we also simulated local transport of $NO_2$ or $NO_x$ along elevation (Fig. 2E). Therefore, we mainly simulated temporal evolution of $NO_2$ or $NO_x$ concentration along the horizontal longitude ($l_x$) and latitude ($l_y$), and elevation ($l_z$) (i.e. a 2-D terrain following system) using Reynolds decomposition (*1, 2*):

$$\frac{\partial C}{\partial t} = -v_{l_x}\frac{\partial C}{\partial l_x} - v_{l_y}\frac{\partial C}{\partial l_y} - v_{l_z}\frac{\partial C}{\partial l_z} + p_{l_x}\frac{\partial^2 C}{\partial^2 l_x} + p_{l_y}\frac{\partial^2 C}{\partial^2 l_y} + p_{l_z}\frac{\partial^2 C}{\partial^2 l_z} + \rho \tag{S2}$$

where $v_{l_x}$ (or $v_{l_y}$ or $v_{l_z}$) is the Reynolds velocity of $NO_2$ or $NO_x$ of the $l_x$ (or $l_y$ or $l_z$) direction of the grid or target location, $\frac{\partial C}{\partial l_x}$ (or $\frac{\partial C}{\partial l_y}$ or $\frac{\partial C}{\partial l_z}$) denotes the advection partial derivative for the corresponding direction, and $p_{l_x}$ (or $p_{l_y}$ or $p_{l_z}$) is the diffusion coefficient for the $l_x$ (or $l_y$ or $l_z$) direction, $\frac{\partial^2 C}{\partial^2 l_x}$ (or $\frac{\partial^2 C}{\partial^2 l_y}$ or $\frac{\partial^2 C}{\partial^2 l_z}$) represent the eddy diffusion terms for the corresponding direction. For the vertical direction of $l_z$, it was naturally restricted to a limited range of elevation observed in the data to account for changes in ground $NO_2$ or $NO_x$ concentrations. The $NO_2$ or $NO_x$ concentration variations over time (Fig. 2) consist of three components, advection, diffusion and the others (transformation, emission and deposition). $\rho$ denotes the variation of air pollutant concentration from emissions, transformation and deposition.

Our joint physics-informed neural network (jPINN) consists of two full residual neural networks that shared the same input, total loss function and joint output of $NO_2$ and $NO_x$ (Fig. S1). One residual neural network (Fig. S1-A) was used to learn the parameters (each direction Reynolds velocity, each direction diffusion coefficient, and $\rho$) to be used in simulation of the continuity equations, and the other residual neural network (Fig. S1-B) was used to predict $NO_2$ and $NO_x$ concentration. Full residual skip connections in two neural networks were used to improve learning efficiency in regression (*3*). The velocity, diffusion coefficient and $\rho$ have been assumed to be affected considerably by meteorology, geography, terrain and vertical atmospheric variables (*2, 4, 5*). In order to avoid complex parameterization in the advection and diffusion simulation, we constructed the specialized neural network to estimate these parameters with possible constraints. There may be differences in network topology and configuration between two neural networks. Through sensitivity analysis, we obtained an optimal network topology for the network of estimating parameters and the network of joint estimating $NO_x$, respectively: the number of nodes in

each encoding or coding layer was 1024→512→320→256→128→96→64→32→16→8 for parameter network and 512→320→256→128→96→64→32→16 for estimation network; the number of nodes in each decoding layer was symmetric with its corresponding coding layer. In addition, the attention layers are added to weigh the influence of different parts of the input on the model's predictions (*6, 7*); the weight normalization (*8*) and non-linear activation functions such as swish and tanh (*9*) layers were added to improve generalizability of the trained neural network.

Encoding fluid physics via PDE residuals

As a differentiable programing paradigm, Tensorflow uses automatic differentiation to optimize the parameters in training and also can be used to solve partial differentiation equation (PDE) problems (*10*). We used Tensorflow as the tool of deep learning for our modeling architecture. Based on Eq. S2, we derived the following PDE residual for $NO_2$ and $NO_x$:

$$e_k \approx e_k' = \frac{\partial C_k'}{\partial t} + v_{l_x}^{(k)} \frac{\partial C_k'}{\partial l_x} + v_{l_y}^{(k)} \frac{\partial C_k'}{\partial l_y} + v_{l_z}^{(k)} \frac{\partial C_k'}{\partial l_z} - p_{l_x}^{(k)} \left( \frac{\partial^2 C_k'}{\partial^2 l_x} + \left(\frac{\partial C_k'}{\partial l_x}\right)^2 \right)$$
$$- p_{l_y}^{(k)} \left( \frac{\partial^2 C_k'}{\partial^2 l_y} + \left(\frac{\partial C_k'}{\partial l_y}\right)^2 \right) - p_{l_z}^{(k)} \left( \frac{\partial^2 C_k'}{\partial^2 l_z} + \left(\frac{\partial C_k'}{\partial l_z}\right)^2 \right) - \rho^{(k)}$$

(S3)

where the dependent variable ($NO_2$ or $NO_x$ concentration), $C_k$ ($k$=1: $NO_2$; $k$=2: $NO_x$) was log-transformed into $C_k' = \log(C_k)$ and so we used $e_k'$ to approximate $e_k$ of the concentration, $C_k$, $x$ and $y$ denote longitude and latitude, respectively, $z$ denotes elevation, $t$ denotes the week index. During model training, minimizing the loss function makes the residuals of $e_1$ and $e_2$ close to 0, enforcing the trained model to conform to the continuity equation as closely as possible. The parameters of Reynolds velocity, diffusion coefficient, and $\rho$ can be learned from the input data (Table S1) of meteorology, emissions, topography and landuse etc. Thus, 14 parameters, $\theta = \left(v_{l_x}^{(1)}, v_{l_y}^{(1)}, v_{l_z}^{(1)}, p_{l_x}^{(1)}, p_{l_y}^{(1)}, p_{l_z}^{(1)}, \rho^{(1)}, v_{l_x}^{(2)}, v_{l_y}^{(2)}, v_{l_z}^{(2)}, p_{l_x}^{(2)}, p_{l_y}^{(2)}, p_{l_z}^{(2)}, \rho^{(2)}\right)$ can be solved using a network (Fig. S1-A).

Encoding thresholds and relationships

To encode the thresholds for $NO_2$ and $NO_x$, we define the following residuals:

$$e_3 = \text{ReLU}(C - maxC) \tag{S4}$$

$$e_4 = \text{ReLU}(C' - maxC') \tag{S5}$$

where $C$ and $C'$ denote $NO_2$ and $NO_x$ concentrations, respectively, $maxC$ denotes maximum allowable $NO_2$ concentration, and $maxC'$ denotes maximum allowable $NO_x$ concentration. When $C > maxC$ (or $C' > maxC'$), $e_3$>0 ($e_4$>0) will lead to an increase in the loss, which in turn back-propagated to optimize the parameters, hence making the loss smaller in optimization.

To encode the relationship between $NO_2$ and $NO_x$ ($NO_2 \leqslant NO_x$), we defined the following residual:

$$e_5 = \text{ReLU}(C - C') \tag{S6}$$

When $C > C'$, $e_5$>0 will lead to an increase in the loss, which in turn back-propagated to optimize the parameters, hence making the loss smaller in optimization.

Loss function and optimization algorithm

According to the continuity equation in Eq. S1 and common-sense knowledge, the residuals of $e_1$-$e_5$ should be hold for all the samples including training, testing and predicting. In addition, the mean square error (MSE) residuals can be defined for the training samples:

$$e_6 = y(C) - C \tag{S7}$$

$$e_7 = y(C') - C' \tag{S8}$$

where $y(C)$ and $y(C')$ denotes observed concentrations of $NO_2$ and $NO_x$, respectively, and $C$ and $C'$ denotes predicted $NO_2$ and $NO_x$ concentrations, respectively. The target variables, concentrations of ground $NO_2$ and $NO_x$, can be solved using a network (Fig. S1-B) in support of the network of parameters.

To maintain the loss function as sufficiently differentiable, the total loss function ($\mathcal{L}$) is defined as the sum of the mean squared residuals multiplied by their respective weights ($\lambda_i$) (*11*):

$$\mathcal{L} = \frac{1}{N}\sum_{i=1}^{5}\sum_{j=1}^{N}\lambda_i e_i(\mathbf{X}_j)^2 + \frac{1}{M}\sum_{i=6}^{7}\sum_{j=1}^{M}\lambda_i e_i(\mathbf{X}_j)^2 \tag{S9}$$

where $N$ denotes the number of all the samples, $M$ denotes the number of training samples, and $X_j$ is the input matrix.

For model training, the input data had three groups of samples: training, regular testing and predicting. The training samples used the measurements of $NO_2$ and $NO_x$

**Algorithm 1:** Joint PINN Learning Algorithm

| | | |
|---|---|---|
| **Input** | : | Input X and Y: $(\mathbf{X}_{tr}, Y_{tr}), (\mathbf{X}_{te}, Y_{te}), (\mathbf{X}_{site}, Y_{site})$ ; |
| | | Mini-batch size: $b$; Number of learning epochs: $n_{epoch}$ ; |
| | | Nodes and acts for parameter net: $\mathcal{L}_p, \mathcal{A}_p$ ; |
| | | Nodes and acts for NOx estimating net: $\mathcal{L}_e, \mathcal{A}_e$ ; |
| | | Thresholds for $NO_2$ and $NO_x$: $Max_1, Max_2$ |
| **Output** | : | Trained model: $\mathcal{M}$ and metrics |
| **Function** | : | Partial derivative by AD: $grad$ ; |
| | | Sub full residual neural network: $frnn$ |
| **Parameter:** | | Sub neural network for parameter estimation: $\mathcal{M}_{para}$ ; |
| | | Sub neural network for $NO_2$ and $NO_x$ estimation: $\mathcal{M}_{est}$ |

1 Initialize the parameters ;
2 $\mathcal{M}_{para} \leftarrow frnn(\mathcal{L}_p, \mathcal{A}_p), \mathcal{M}_{est} \leftarrow frnn(\mathcal{L}_e, \mathcal{A}_e)$;
3 Obtain $\mathcal{M}$ by connecting $\mathcal{M}_{para}$ and $\mathcal{M}_{est}$ ;
4 $l_{mini} \leftarrow \text{len}(X_{tr})/b$ ;
5 **for** $i = 1 \cdots n_{epoch}$ **do**
6      Shuffle the input data index: $\text{index}_{te}, \text{index}_{tr}, \text{index}_{site}$ ;
7      $\mathcal{T}_{\mathcal{L}} \leftarrow 0$ ;
8      **for** $j = 1 \cdots l_{mini}$ **do**
9          Calculate the minibatch index: $m_{tr}, m_{te}, m_{site}$ ;
10          $\mathbf{X}_{all} \leftarrow \text{concate}([\mathbf{X}_{tr}[m_{tr}], \mathbf{X}_{te}[m_{te}], \mathbf{X}_{site}[m_{site}]])$ ;
11          $\hat{Y} \leftarrow \mathcal{M}_{est}(\mathbf{X}_{all}), \theta \leftarrow \mathcal{M}_{para}(\mathbf{X}_{all})$ ;
12          $\partial \hat{Y} / \partial t \leftarrow grad(\hat{Y}, t)$;
13          $\partial \hat{Y} / \partial x \leftarrow grad(\hat{Y}, x), \partial^2 \hat{Y} / \partial^2 x \leftarrow grad(\partial \hat{Y} / \partial x, x)$ ;
14          $\partial \hat{Y} / \partial y \leftarrow grad(\hat{Y}, y), \partial^2 \hat{Y} / \partial^2 y \leftarrow grad(\partial \hat{Y} / \partial y, y)$ ;
15          $\partial \hat{Y} / \partial z \leftarrow grad(\hat{Y}, z), \partial^2 \hat{Y} / \partial^2 z \leftarrow grad(\partial \hat{Y} / \partial z, z)$ ;
16          $e_1 = \partial \hat{Y}_1 / \partial t + \theta_1 * \partial \hat{Y}_1 / \partial x + \theta_2 * \partial \hat{Y}_1 / \partial y + \theta_3 * \partial \hat{Y}_1 / \partial z$;
17          $e_1 = e_1 - \theta_4(\partial^2 \hat{Y}_1 / \partial^2 x + (\partial \hat{Y}_1 / \partial x)^2) - \theta_5(\partial^2 \hat{Y}_1 / \partial^2 y + (\partial \hat{Y}_1 / \partial y)^2)$;
18          $e_1 \leftarrow e_1 - \theta_6(\partial^2 \hat{Y}_1 / \partial^2 z + (\partial \hat{Y}_1 / \partial z)^2) - \theta_7$;
19          $e_2 = \partial \hat{Y}_2 / \partial t + \theta_8 * \partial \hat{Y}_2 / \partial x + \theta_9 * \partial \hat{Y}_2 / \partial y + \theta_{10} * \partial \hat{Y}_2 / \partial z$;
20          $e_2 = e_2 - \theta_{11}(\partial^2 \hat{Y}_2 / \partial^2 x + (\partial \hat{Y}_2 / \partial x)^2) - \theta_{12}(\partial^2 \hat{Y}_2 / \partial^2 y + (\partial \hat{Y}_2 / \partial y)^2)$;
21          $e_2 \leftarrow e_2 - \theta_{13}(\partial^2 \hat{Y}_2 / \partial^2 z + (\partial \hat{Y}_2 / \partial z)^2) - \theta_{14}$;
22          $e_3 \leftarrow \text{ReLU}(\hat{Y}_1 - Max_1), e_4 \leftarrow \text{ReLU}(\hat{Y}_2 - Max_2)$ ;
23          $e_5 \leftarrow \text{ReLU}(\hat{Y}_1 - \hat{Y}_2)$ ;
24          $\hat{Y}_{tr} \leftarrow \mathcal{M}_{est}(\mathbf{X}_{tr}), e_6 \leftarrow Y_{tr1} - \hat{Y}_{tr1}, e_7 \leftarrow Y_{tr2} - \hat{Y}_{tr2}$ ;
25          $\mathcal{L} \leftarrow \text{mean}(e_1^2) + \text{mean}(e_2^2) + \text{mean}(e_3^2) + \text{mean}(e_4^2)$ ;
26          $\mathcal{L} \leftarrow \mathcal{L} + \text{mean}(e_5^2) + \text{mean}(e_6^2) + \text{mean}(e_7^2)$;
27          Optimize model parameters based on $\mathcal{L}$ and gradients ;
28          $\mathcal{T}_{\mathcal{L}} \leftarrow \mathcal{T}_{\mathcal{L}} + \mathcal{L}$ ;
29      **end for**
30      $\mathcal{T}_{\mathcal{L}} \leftarrow \mathcal{T}_{\mathcal{L}} / l_{mini}$ ;
31      Calculate raw $R^2$ and RMSE of training, test and site-based test;
32 **end for**
33 Return $\mathcal{M}, \mathcal{T}_{\mathcal{L}}, R^2, \text{RMSE}$

concentrations to train the model through the MSE residuals of $e_6$ and $e_7$, whereas the regular test and prediction samples did not use $NO_2$ and $NO_x$ measurements ($Y'$) in training but used them for testing and/or validation. Since all the samples need to satisfy the continuity equation, threshold and relation constraints, the input data (**X**) for training and regular testing were used to train the model through the squared residuals of $e_1$-$e_5$. For ensemble predictions, we used all the samples (including training and testing) as unsupervised samples to train the models to make them more regular, conforming to the continuity equation and physical constraints. The learning algorithm is briefly described in Algorithm 1. The sensitivity analysis was conducted to find optimal solutions for the hyperparameters including mini-batch size (1666), number of epochs (160), learning rate (0.01), beta (0.09), and epsilon (1e-3). The clip norm was set as 1 to fix potential exploding gradients. We mainly used the combination of non-linear activation of swish and elu plus linear activation in the network architecture. Sensitivity analysis showed that nonlinear activations of pure swish, tanh, or sigma, or their mixtures may lead to vanishing gradients and slow convergence, while the appropriate combination of nonlinear and linear activations (e.g., swish + linear (last layer) activations for network for parameter estimation, and swish + elu activations for network for concentration estimation) achieved the best results.

Uncertainty estimation

The nonparametric bootstrap method was used to assess the uncertainty of our predictions because it performs as well as parametric uncertainty and works for the non-normal and nonlinear data (*12*). As samples from 63.2% of the sampling locations were used to train the model, half of the remaining samples were used for regular testing and the other half were used for site-based testing, as presented in Fig 3.

In total, by bootstrapping, we obtained 150 models to make 150 predictions, and then the mean of the prediction distribution can be estimated:

$$\hat{\mu}_{s,t} = \frac{1}{B} \sum_{b=1}^{B} \hat{y}_{s,t}^{b} \tag{S10}$$

where $\hat{y}_{s,t}^{b}$ is the estimate at spatial location, $s$ and time point, $t$ through bootstrap $b$.

The model variance can be estimated using the bootstrapping predictions:

$$\hat{\eta}_{s,t}^{v,b} \approx \hat{\mu}_{s,t} - \hat{y}_{s,t}^{b} \quad b = 1,...,B \tag{S11}$$

where $\eta_{s,t}^{v,b}$ denotes the model variance of location $s$ and time point $t$ for bootstrap $b$.

The model bias and sample noise can be estimated from training error, regular testing error and site-based independent testing error. These errors are defined as the difference between the observed and predicted concentrations of $NO_2$ and $NO_x$ for the training, regular testing and site-based testing samples of each bootstrap, respectively. The training and regular testing errors may be overly optimistic and the site-based testing errors may be overly pessimistic. According to the sampling proportion in bootstrap, the balance between these errors has been suggested as "0.632 + bootstrap estimate" (*13*) for deriving model bias and sample noise (*12*). Here, we derived the estimation of model bias and sampling noise according to the sampling proportion:

$$b_{s,t} = \eta_{s,t}^{b} + \varepsilon_{s,t}^{n} \approx \left(1 - \hat{w}_{te} - \hat{w}_{site}\right) \times \hat{\varepsilon}_{train} + \hat{w}_{te} \times \hat{\varepsilon}_{te} + \hat{w}_{site} \times \hat{\varepsilon}_{site} \tag{S12}$$

where $\hat{\varepsilon}_{train}$ denotes estimated training error, $\hat{\varepsilon}_{te}$ denotes estimated regular testing error, $\hat{\varepsilon}_{site}$ denotes estimated site-based testing error, $\eta_{s,t}^{b}$ denotes the model bias, $\varepsilon_{s,t}^{n}$ denotes random noise, $\hat{w}_{te}$ is the weight for regular testing error, and $\hat{w}_{site}$ is the weight for site-based testing error.

$$\hat{w}_{te} = \frac{0.632}{\left(1 - 0.184 \hat{R}_{te}\right)} \tag{S13}$$

$$\hat{w}_{site} = \frac{0.632}{\left(1 - 0.184 \hat{R}_{site}\right)} \tag{S14}$$

where $\hat{R}_{te}$ denotes the relative overfitting rate for regular testing, defined as $\hat{R}_{te} = \frac{(\hat{\varepsilon}_{te} - \hat{\varepsilon}_{train})}{(\hat{\gamma} - \hat{\varepsilon}_{train})}$, $\hat{R}_{site}$ denotes the relative overfitting rate for site-based testing, defined as $\hat{R}_{site} = \frac{(\hat{\varepsilon}_{site} - \hat{\varepsilon}_{train})}{(\hat{\gamma} - \hat{\varepsilon}_{train})}$, and $\hat{\gamma}$ is the no-information error rate, defined as $\frac{1}{n^2} \sum_{s,t} \sum_{s',t'} \left(y_{s,t} - \hat{y}_{s',t'}\right)^2$ (*12*).

According to Eq. 1, S34 and S35, the combined set can be used to approximate the distribution of prediction errors. In addition, the predicted concentration was classified as 8 levels (Fig. 3) and the samples of corresponding level were used to derive the uncertainty estimate.

$$E_l = \left\{ \eta_{s_1,t_1}^{v,b} + o_{s_2,t_2}^{l} \mid b < B, s_2 < N_l, t_2 < M_l \right\} \quad \text{(S15)}$$

where $l$ is the level of predicted concentration, $s_1$ and $t_1$ are the location index and time index of the predicted sample, respectively, $o_{s_2,t_2}^{l}$ is the sample value of bias and sampling noise for level $l$, $N_l$ is the number of locations for level $l$, and $M_l$ is the number of time points for level $l$.

Subsequently, the standard deviations of the predictions were estimated and the prediction intervals were obtained by the prediction offset by the percentiles of $\left(100 \cdot \frac{\alpha}{2}\right)\%$ and $\left(100 \cdot \left(1-\frac{\alpha}{2}\right)\right)\%$ of the set of Eq. 38 ($\alpha$ is the significance level; $\alpha$=0.05 for this case study).

Study region

The study region covers the entire state of California (CA) (Fig. S2), located between about -124°65' and -114°13' west to east longitude and between 32°51' and 42°01' north to south latitude. With the land of approximately 100 million acres, CA has heterogenous topography (city, coastline, desert, mountain, snow cover etc.), multiple emission sources for air pollutants (anthropogenic sources: fossil fuel exhaust, agriculture and biomass etc.; natural sources: wildfires, dusts and oceans etc.), and complex atmospheric physiochemical processes (*14*).

Over the past 30 years, California has strictly enforced air quality regulation and ambient levels of air pollutants have dropped significantly (*15*), but many parts of the state still have serious air quality problems. In fact, Los Angeles County remains one of the most polluted areas in the United States (*16*). Nitrogen oxides ($NO_x$) are a group of gases regulated by the government – with many important sources in CA.

Dataset

The study duration spanned from Jan. 2004 to Dec. 2020. We obtained the measurement data of $NO_2$ and $NO_x$ from two sources: (1) the US Environmental Protection Agency Air Quality System (AQS), which provides long-term hourly measurements from 193 routine monitoring sites operated by the California Air Resources Board (CARB) and other governmental agencies (2004-2020), and (2) three sampling campaigns where weekly or biweekly intensive field measurements were gathered by the University of Southern California (1104 sampling locations in 12 southern California communities, 2005-2009) (*17, 18*), University of California, Los Angeles (184 sampling locations in Los Angeles County, 2006-2007), and University of California, Irvine (32 sampling locations in south Los Angeles County and Orange County, 2009), respectively. All AQS monitors used the Federal Reference Method (FRM), and the weekly and biweekly field measurements used passive diffusion-based Ogawa samplers to measure $NO_2$ and $NO_x$ concentrations. The passive measurements were systematically adjusted to be consistent with the continuous FRM $NO_x$ and $NO_2$ measurements. We then merged hourly and biweekly concentrations to weekly averages (average concentrations in a weekly time series starting from 01/01/2014 and ending on 12/31/2020). Part of the data was also used and described further in our previous studies (*19, 20*).

As a semi-parametric approach, temporal basis functions have been usually used to simulate the leading temporal modes of variability of air pollutants for a region where the samples are obtained at routine measurement sites (*21*). From the data of long time series at routine monitoring sites, we extracted the first and second temporal basis functions to represent seasonal variations.

On-road traffic, including light- and heavy-duty vehicles, are a primary emissions source of $NO_2$ and $NO_x$. We developed model input variables to reflect the contribution of on-road traffic to ambient pollution concentration. To develop all variables, we used HERE Technology road network data (2016-2019) (https://www.here.com, accessed in 2022) to represent roadway geometry, which classifies roadways into 5 functional classes based on the speed and volume of traffic. Total traffic volume and speed on each roadway link were obtained from Streetlytics™ data by Bentley Systems, Inc. (https://www.bentley.com/software/streetlytics, accessed in 2022 ), and heavy duty volume was obtained from the Traffic census program of the California Transportation Department (https://dot.ca.gov/programs/traffic-operations/census, access in 2022). During the SARS-CoV-2 pandemic, total traffic volumes were scaled at the county level to reflect changes in total traffic volume reported in the Caltrans Performance Measurement System (PeMS) system (https://dot.ca.gov/programs/traffic-operations/mpr/pems-source,

accessed in 2023). A statewide average scaling factor was used to adjust heavy duty traffic volumes for the COVID period.

Variables reflecting roadway geometry included distance to major roadway and intersection density. Distance to major (functional class 1 or 2) roadway (m) was calculated as the euclidian distance of each location to the nearest major roadway. Intersection density data were calculated as the density of intersections in 300 and 5000 meter buffers. Intersections were divided into 4 distinct classes based on the type of roadway involved in the intersection. Intersections involving only major roadways (functional class 3 and 4 roads) were assigned as class 1 intersections. Intersections involving major and minor roads (functional class 5 roads) were assigned as class 2 intersections. Intersections involving only minor roads were assigned as class 3 intersections. Any intersections involving roads for which through traffic is prohibited were assigned as class 4 intersections, which based on inspection, primarily represent parking lots and some minor residential locations. No intersections with freeways (functional class 1 and 2 roads) were used, as on ramps and off ramps are included in lower road classes.

Variables reflecting traffic activity and emissions included traffic load and ambient concentration of traffic-related $NO_x$. Traffic load was calculated as annual average daily vehicle meters within 300 m and 5,000 m of each location. Traffic load was calculated separately for light and heavy duty vehicles and for freeways (functional class 1 and 2 roads) and other roads (functional class 3-5 roads). Ambient concentration of traffic-related $NO_x$ (ppb) was calculated using the CALINE4 line source dispersion model with contributions separated by freeways and other roads (*22, 23*).

Land-use variables can capture emission sources and sinks of $NO_2$ and $NO_x$. From the National Land Cover Database (NCLD) (https://www.mrlc.gov) at the 30m resolution (*24*), we generated annual cover of each land cover classes (e.g., open water, developed with low intensity, developed with high intensity, barren land, grassland and cultivated crops etc.), we calculated the proportion of cover of each class within 1 km grid cell for each available year (2013, 2016, 2018 and 2020) and linearly interpolated to annual values for unavailable years of 2013-2020. We excluded the land cover types that were not as frequently observed in CA (<1%). Using a similar interpolation, we also extracted the surface impervious layer that represents urban impervious surfaces as a percentage of developed surfaces over every 30m pixel for 2013-2020.

Meteorology plays very important roles in forming, dispersion and transport of $NO_2$ and $NO_x$ (*25*). We obtained meteorological parameters from daily high spatial resolution (~4km, 1/24 degree) surface meteorological data from the gridMET of contiguous US (http://www.climatologylab.org/gridmet.html) (*26*). These parameters included daily minimum air temperature (°C), maximum air temperature (°C), wind speed (meters/second, m/s), specific humidity (grams of vapor per kilogram of air, g/kg), daily mean downward shortwave radiation (watt/meter$^2$, w/m$^2$) and accumulated precipitation (millimeters of rain per meter$^2$ in 1 h, mm/m$^2$). We generated weekly averages from daily values of the meteorological variables. Bilinear resampling was used to convert the meteorological data into our target UTM Zone 11 projection.

Although our PINN can simulate advection and diffusion of $NO_2$ or $NO_x$ on the ground, it has no internal mechanism to model photochemical reactions. Thus, our input includes reanalysis data to account for impacts from background atmospheric chemical processes and transformations, although such reanalysis data has a coarse spatial resolution. We utilized the data from the Multi-Decadal Nitrogen Dioxide and Derived Products from Satellites (MINDS) project, accessible at https://disc.gsfc.nasa.gov/datasets/OMI_MINDS_NO2d_1.1/summary?keywords=MINDS. The primary objective of MINDS is to adapt operational algorithms from the Ozone Monitoring Instrument (OMI) to various satellite instruments, enabling the creation of consistent Level 2 (L2) and Level 3 (L3) nitrogen dioxide ($NO_2$) columns. Additionally, MINDS generates value-added Level 4 (L4) surface $NO_2$ concentrations and $NO_x$ emissions data products, taking into account instrumental differences in a systematic manner (*27*). The instruments utilized in the MINDS project encompass a range of satellite sensors, namely the Global Ozone Monitoring Experiment (GOME) spanning from 1996 to 2011, SCanning Imaging Absorption spectroMeter for Atmospheric CHartographY (SCIAMACHY) from 2002 to 2012, Ozone Monitoring Instrument (OMI) from 2004 to the present, GOME-2 from 2007 to the present, and TROPOspheric Monitoring Instrument (TROPOMI) from 2018 to the present. These instruments collectively contribute to the comprehensive data collection and analysis within the MINDS project. From the bottom layer diagnostics of MINDS, we selected 9 related variables such as nitric oxide, nitrogen dioxide, nitrate mass mixing ratio etc.; from the single-level diagnostics, we selected 13 variables including wind vectors, specific humidity, temperatures at different altitudes, and total ozone etc.

In addition, based on the MERRA-2 wind speeds at different altitudes (https://disc.gsfc.nasa.gov/datasets/M2I1NXASM_5.12.4/summary), we derived indicators of vertical stagnation and wind sheer/mechanical mixing as follows:

$$w_{stag} = \left(\sqrt{u_{50}^2 + v_{50}^2} - \sqrt{u_{10}^2 + v_{10}^2}\right) \quad (16)$$

$$w_{mix} = \left(\sqrt{u_{10}^2 + v_{10}^2} - \sqrt{u_{2}^2 + v_{2}^2}\right) \quad (17)$$

where $w_{stag}$ is the indicator of stagnation, $w_{mix}$ is wind sheer/mechanical mixing, $u_{10}$ is 10-meter eastward wind speed, $v_{10}$ is 10-meter northward wind speed, $u_2$ is 2-meter eastward wind speed, $v_2$ is 2-meter northward wind speed, $u_{10}$ is 50-meter eastward wind speed, and $v_{50}$ is 50-meter northward wind speed.

From remotely sensed satellites, we obtained NDVI. We calculated monthly average NDVI using 16 day MODIS NDVI average values from NASA's Aqua and Terra satellite (MOD13A2 V6 and MCD13A3 V6).

Considering the potential fuel burning from cooking sources for $NO_x$ emissions can be significant especially in urban areas, we extracted shortest distances from fast food facilities and restaurants, respectively, whose locations were accessed through OpenStreet (https://www.openstreet.com, accessed in 2022). We also extracted kernel density with the buffer of the 30 km radius for fast food facilities and restaurants respectively.

We also obtained 30 m resolution elevation from GoogleMaps API. We calculated block group population in 300 m buffers based on the 2000, 2010 census block data in ArcGIS and linearly interpolated or extrapolated annual population density for 2013−2020. All the covariates used in the model are summarized in Table S1.

Model training and comparison

In total, we obtained 81,929 $NO_2$ and $NO_x$ samples of weekly averages over hourly measurements from January 2004 to December 2020 (17 years) from 1342 sampling locations. These locations included 148 routine monitoring sites of the U.S. Environmental Protection Agency Air Quality System (AQS), and 1194 sampling locations of special campaigns (1104: The University of Southern California; 184: University of California Los Angeles; 32 University of California Irvine). Of these sampling locations, 63.26% were selected for training and regular testing to get the supervised samples in training, 18.4% were selected to get the unsupervised samples in training and regular testing samples, and the rest 18.4% were used to get the site-based independent testing samples. For the 81.6% sampling locations for training and regular testing, the data were sampled in a mixed way of spatial and temporal dimensions with county and season as a combinational stratifying factor, and 80% of these samples were used for training and the resting 20% were used for regular testing. The low-concentration or high-concentration samples in the training data were oversampled by approximately 20%.

The same training samples were used to learn the models of all representative models plus our joint PINNs. For our joint PINNs (Fig. 2), the same training samples were used in optimization by the RSE residuals ($e_6$-$e_7$), but the samples including training, regular testing and site-based testing were used in optimization by the physical residuals (PDE: $e_1$-$e_2$; constraint and relationship: $e_3$-$e_5$) since we assumed that all the samples must satisfy the physical principles. Sensitivity analysis was conducted to validate the models only using training samples for RSE residuals, and training and regular testing samples for physical residuals; the site-based testing samples were just used for testing (no use for training the model including physical residuals by $e_1$-$e_5$). The results (Fig. S4 for $R^2$ and Fig. S5 for RMSE) show very similar metrics (mean $R^2$: 0.95-0.96) of the models trained with no site-based samples with those trained using all the samples with a very small difference. The unsupervised samples can help the trained models to regular their boundary points and the missing of the site-based samples might cause such a small difference. The sensitivity analysis further strengthened the reliability and generalizability of our joint PINNs.

**Fig. S1.**

**Modeling architecture of physics-informed residual neural networks for joint estimates of $NO_2$ and $NO_x$. A,** full residual neural network for estimation of the parameters for $NO_2$ and $NO_x$ continuity equations, including each direction Reynolds velocity, diffusion coefficient and $\rho$ (14 parameters in total). **B,** full residual neural network for estimation of $NO_2$ and $NO_x$ concentrations. **C,** Construction of residuals and total loss function, including PDE residuals and threshold residuals of $NO_2$ and $NO_x$, their relationship residuals, and their RMSE residuals. $N$ denotes the number of all the samples, $M$ denotes the number of the training samples, and $\lambda_i$ denotes the weight for each residual. The vertical rectangular orange bar represents the batch normalization layer to be added, the vertical rectangular blue bar represents the activation layer to be added, and the dashed two-dotted line represents the residual connection from the encoding layer to its corresponding decoding layer or around the attention layer.

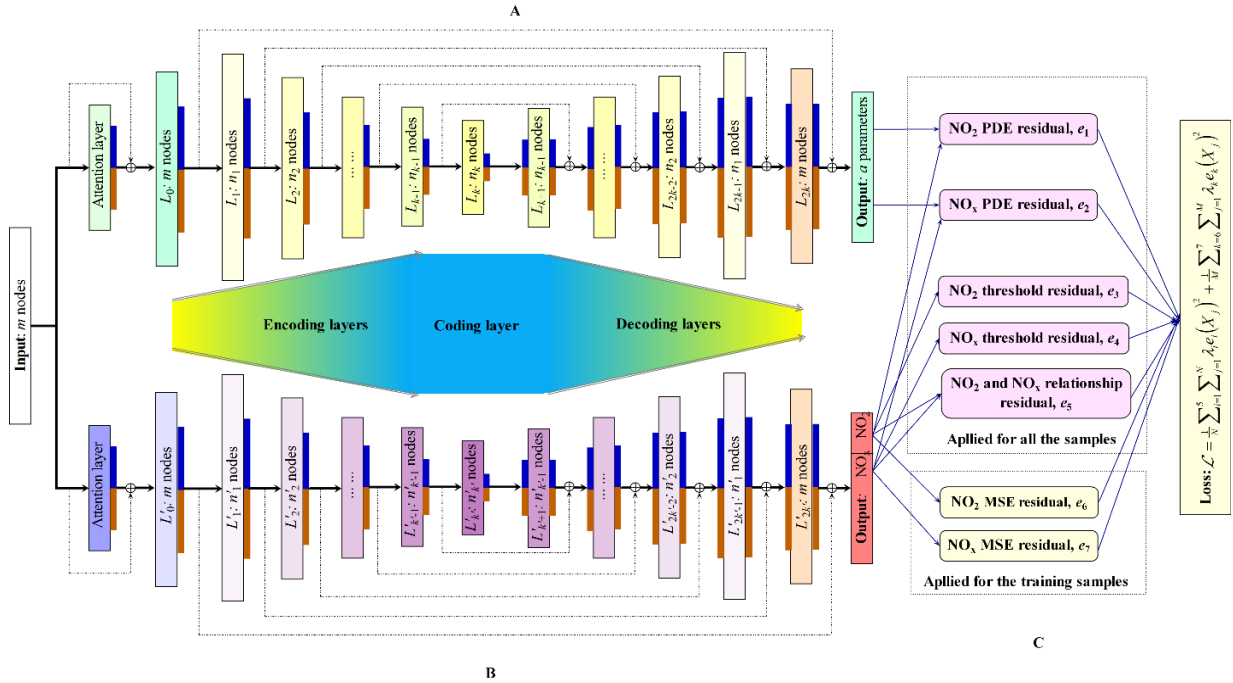

**Fig. S2.**
**Uncertainty analysis of estimated NO₂ and NOₓ concentrations.** (**A**) The process of estimating the bias and variance by bagging of base physics-informed deep learning models. (**B**) The bias estimation by the estimated concentration level.

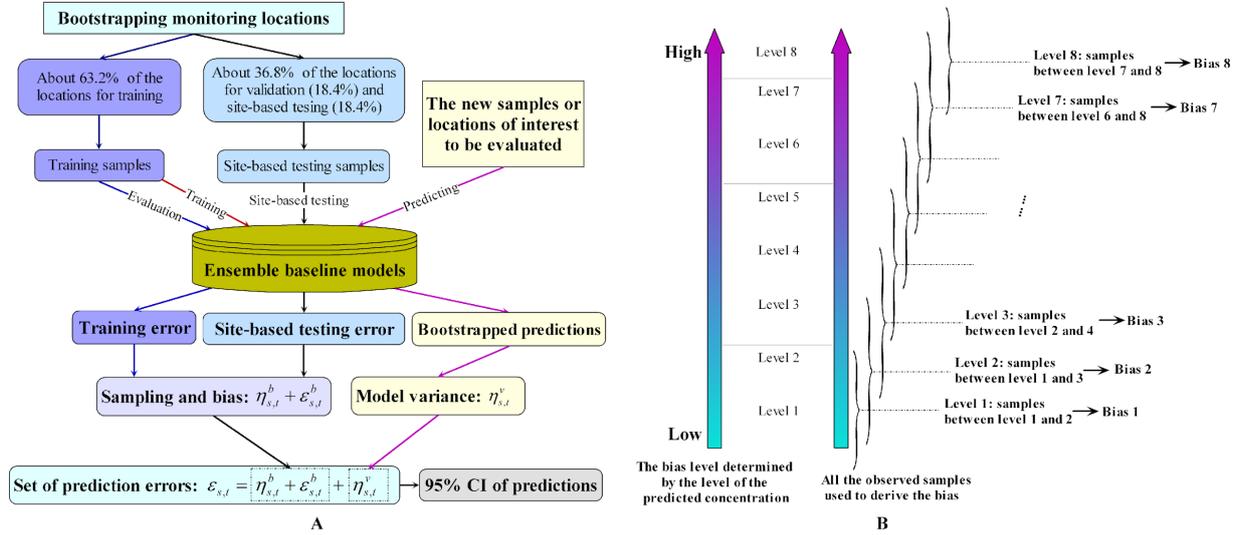

**Fig. S3.**
**Study region with AQS monitoring sites and field sampling locations.**

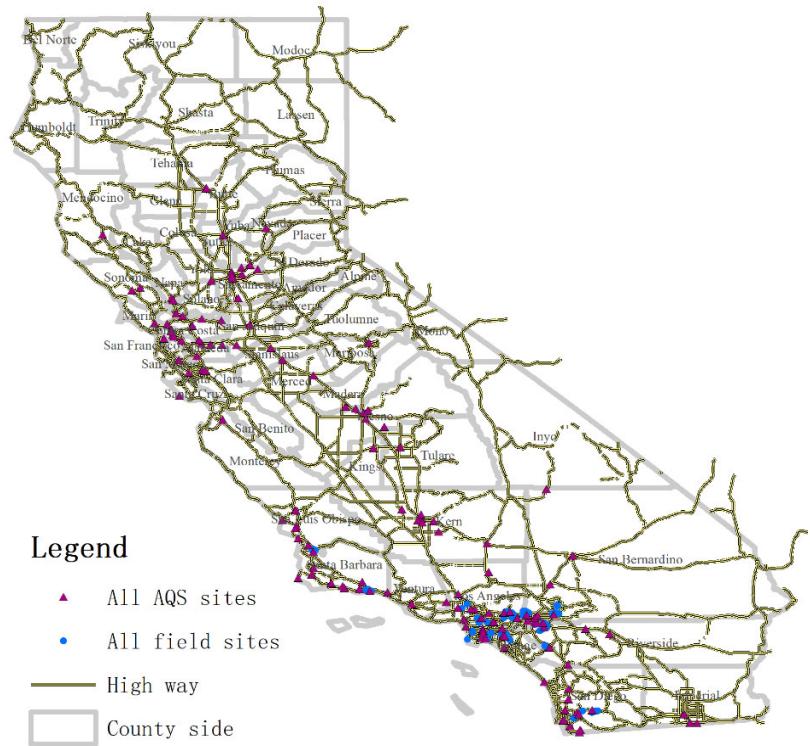

**Fig. S4.**

**Comparison in training, regular testing and site-based testing R² between jPINN with all the samples used to train physical-related residuals (red) and jPINN with only training and regular testing used to train physical-related residuals (green).** Yellow plus indicates the mean of the corresponding group.

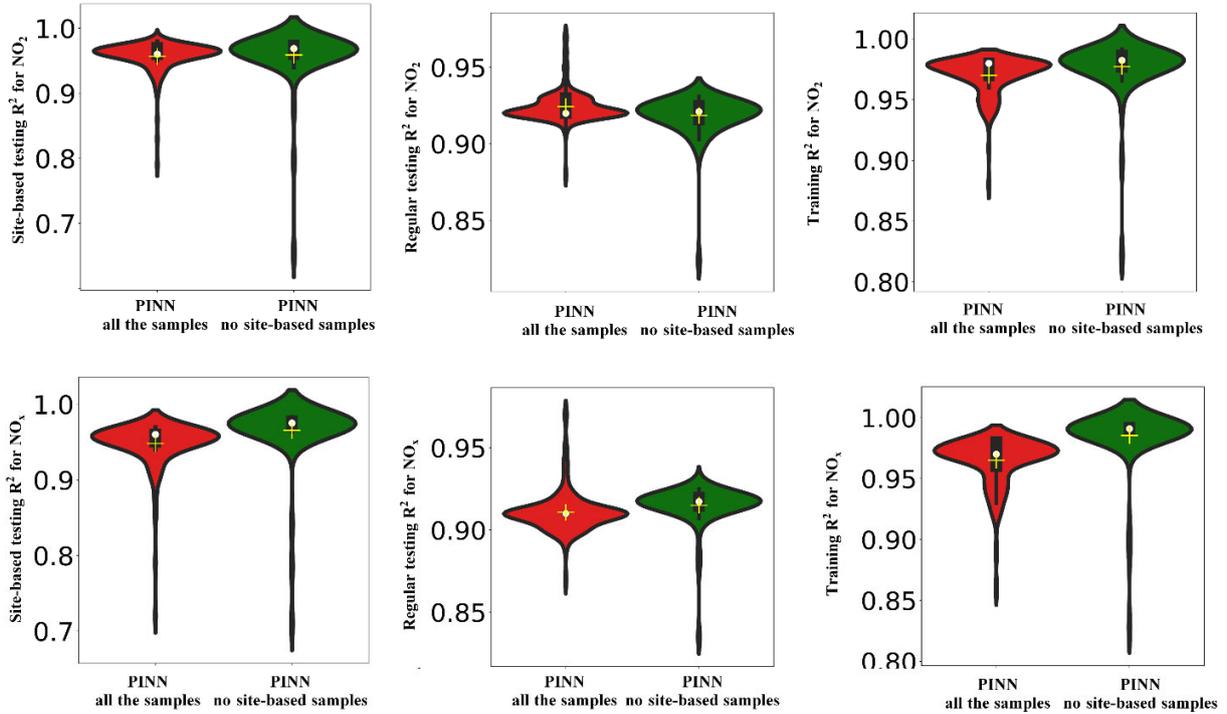

**Fig. S5.**

**Comparison in training, regular testing and site-based testing RMSE (ppb) between jPINN with all the samples used to train physical-related residuals (red) and jPINN with only training and regular testing used to train physical-related residuals (green).** Yellow plus indicates the mean of the corresponding group.

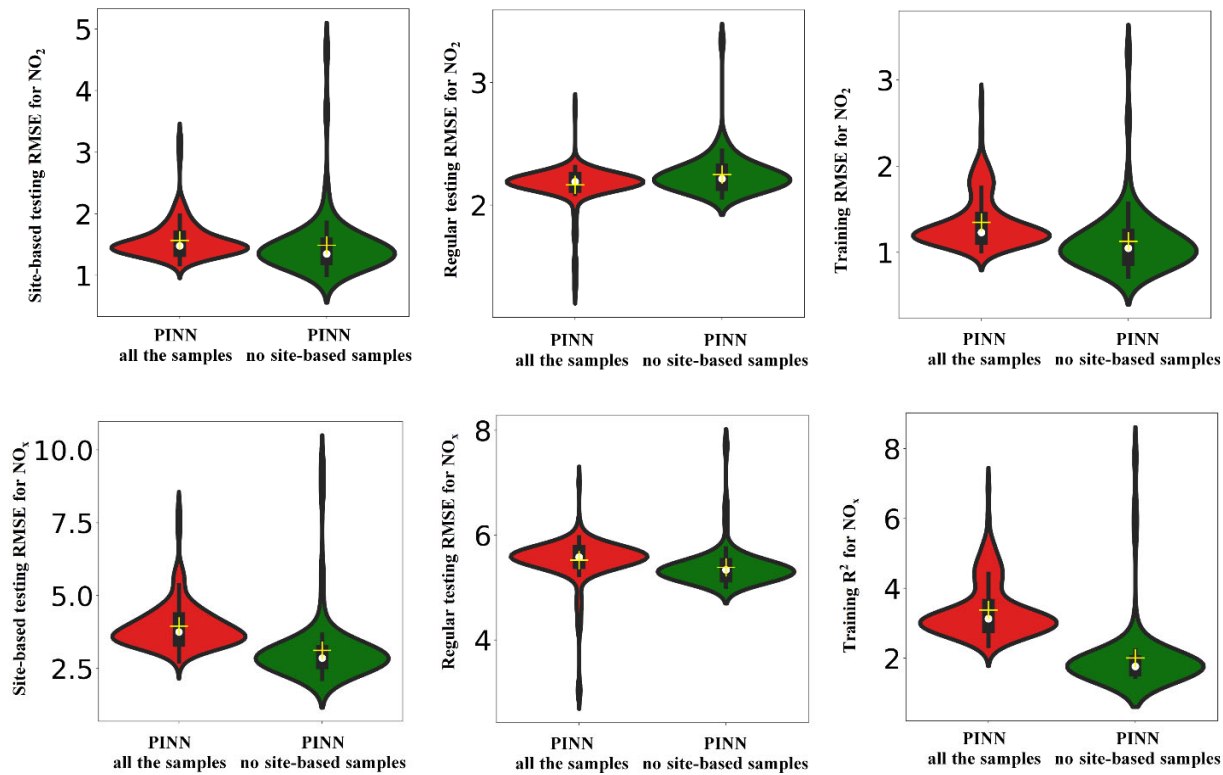

**Fig. S6.**

Violins of site-based independent testing, regular testing and training RMSEs for six representative methods.

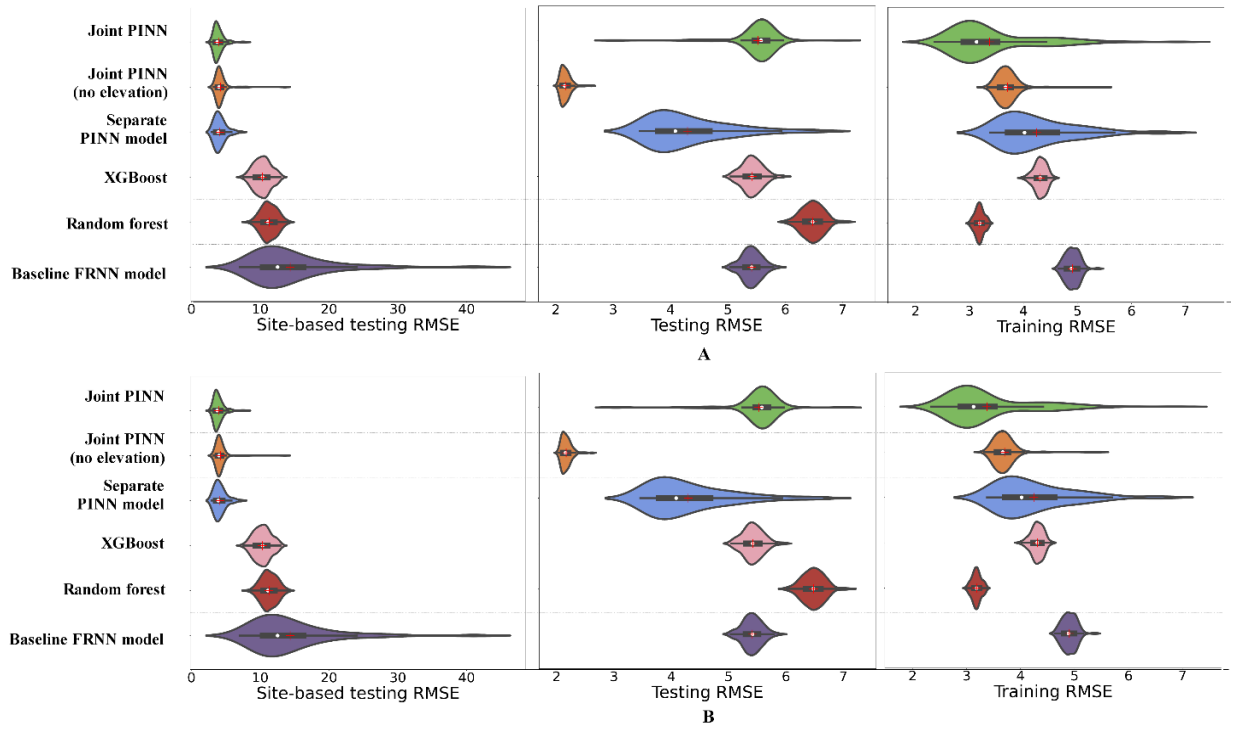

**Fig. S7.**
**Violin plots for the ratios of model variance to generalization error.**

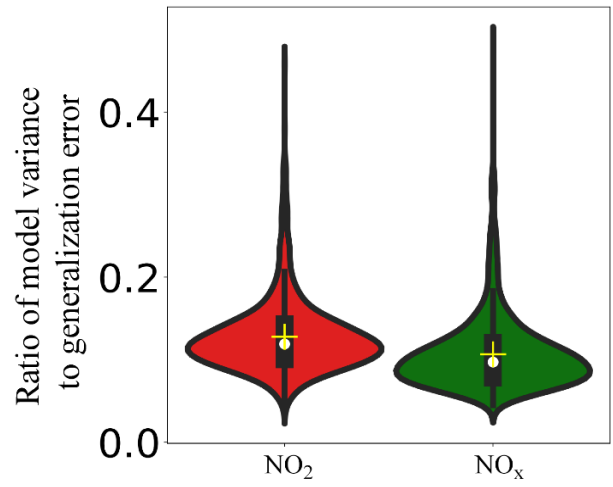

**Fig. S8.**

**Violins of the ratios of $NO_2$ to $NO_x$ for the bagging predicted concentrations for site-based testing samples.**
FRNN: baseline full residual neural network; Joint PINN: Joint Physics-informed neural network; red horizontal line: 1-1 reference line for the ratio of $NO_2$ to $NO_x$.

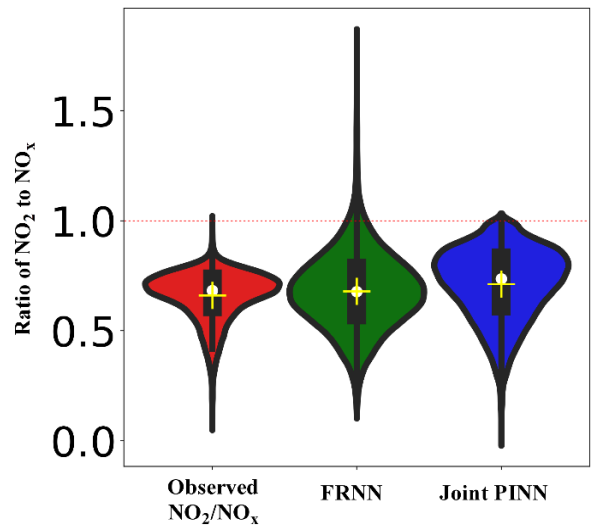

**Fig. S9.**
**Time Series of weekly regular testing (A and C) and site-based independent testing $R^2$ (B and D) for $NO_2$ (A and B) and $NO_x$ (C and D).**

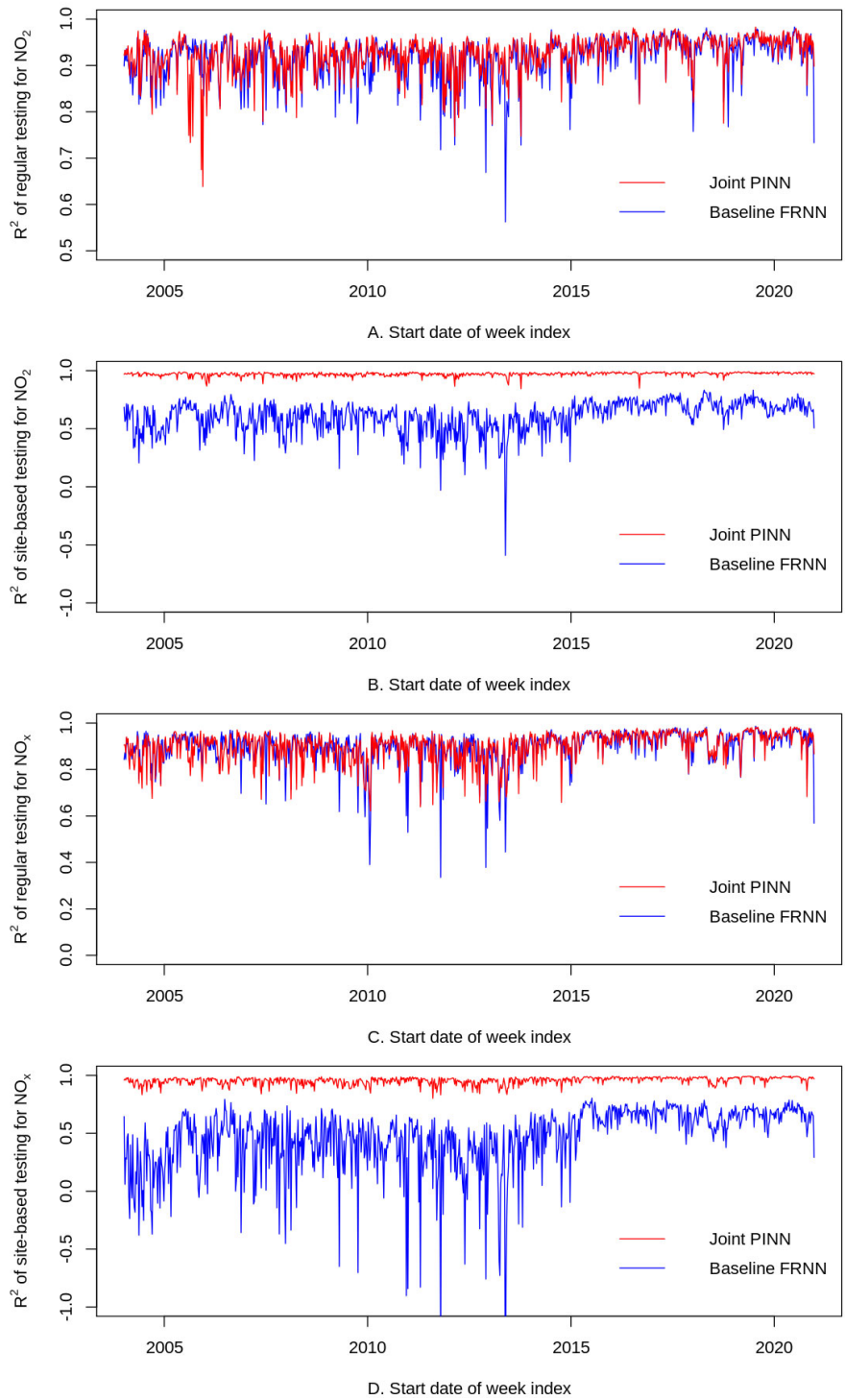

**Fig. S10**

**Time Series of weekly regular testing (A and C) and site-based independent testing RMSE (B and D) for NO₂ (A and B) and NOₓ (C and D).**

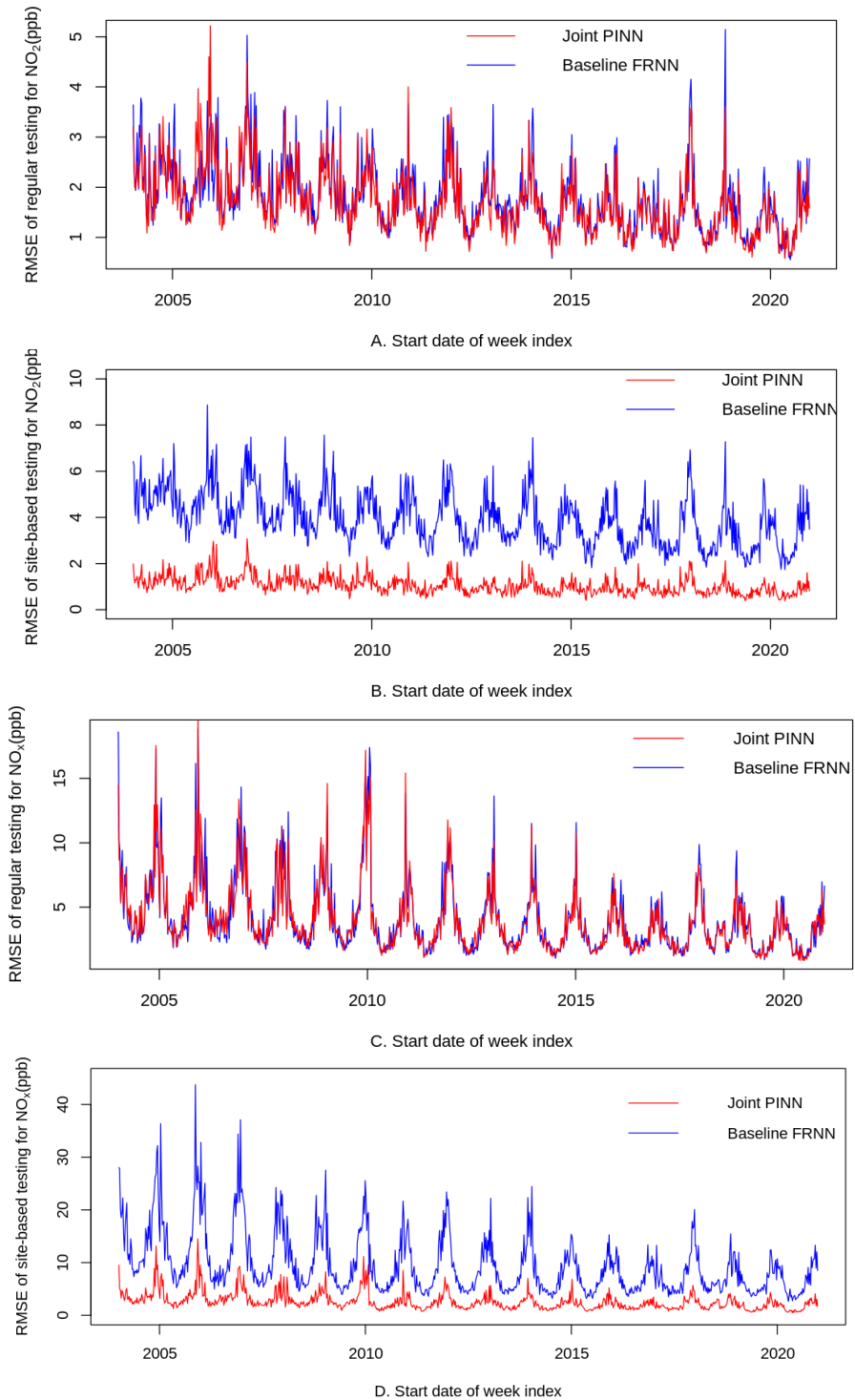

**Fig. S11.**
**Comparison of predicted time series of $NO_2$ and $NO_x$ between baseline FRNNs and jPINNs for a representative AQS site.** The site id was 60850006.

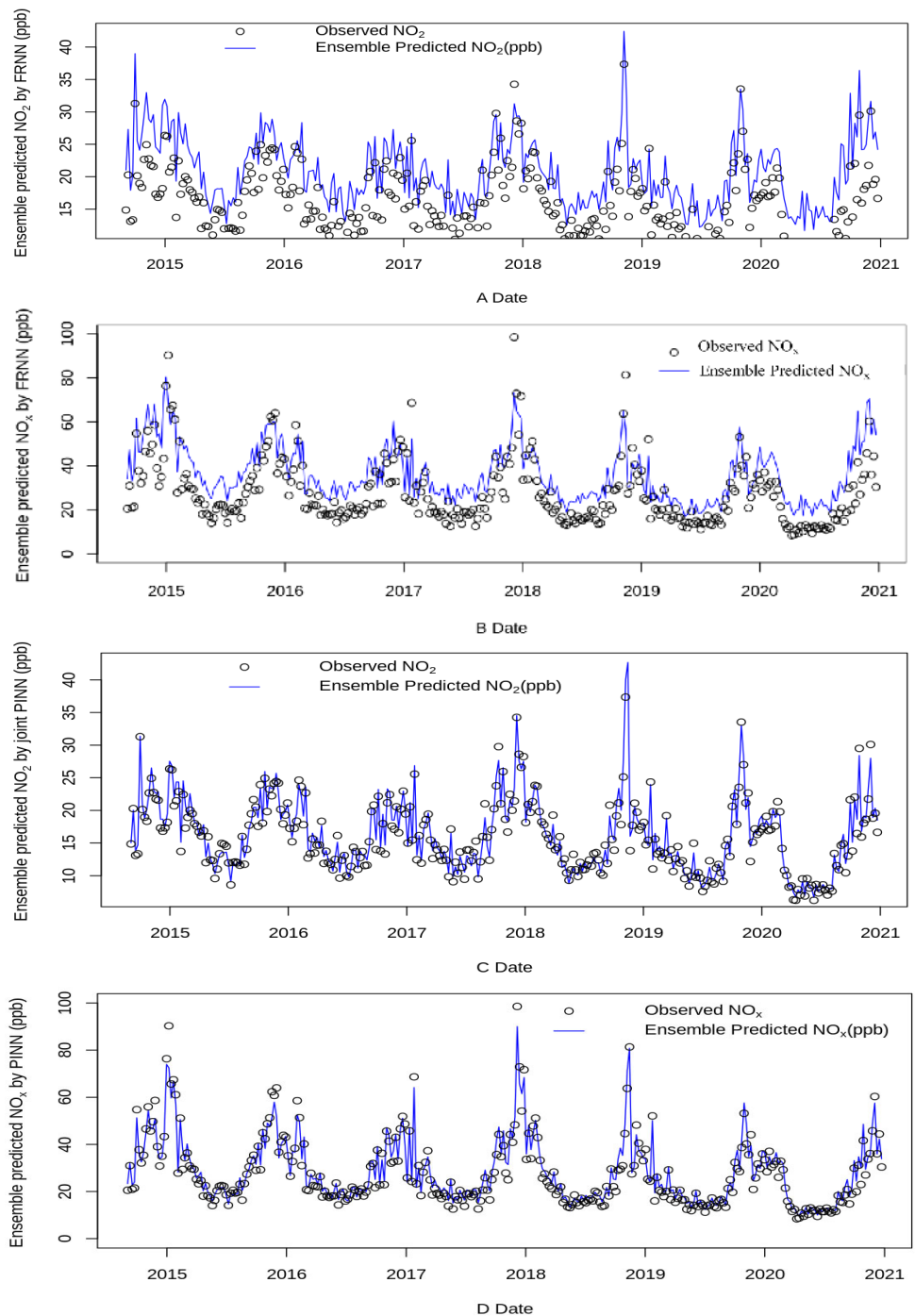

**Fig. S12**
**Sensitivity analysis of the jPINNs for NO₂ to missing MINDS variables and land-use covariates.**

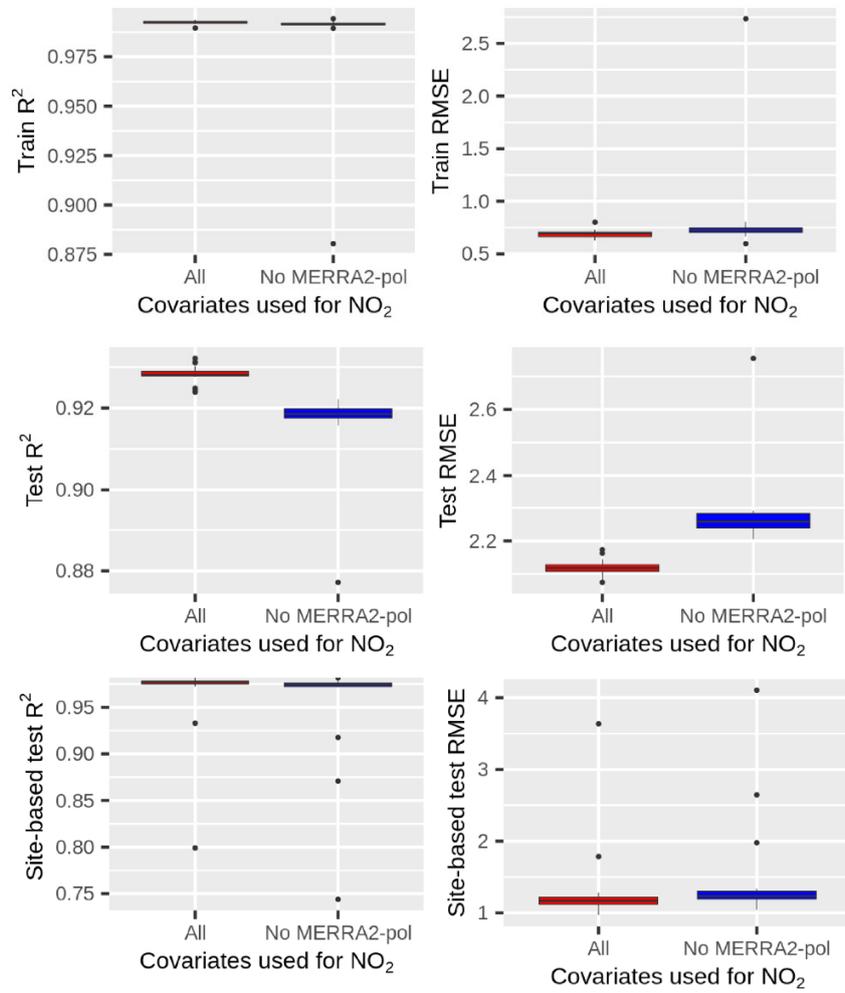

**Fig. S13**
**Sensitivity analysis of the jPINNs for NO$_x$ to missing MINDS variables and land-use covariates.**

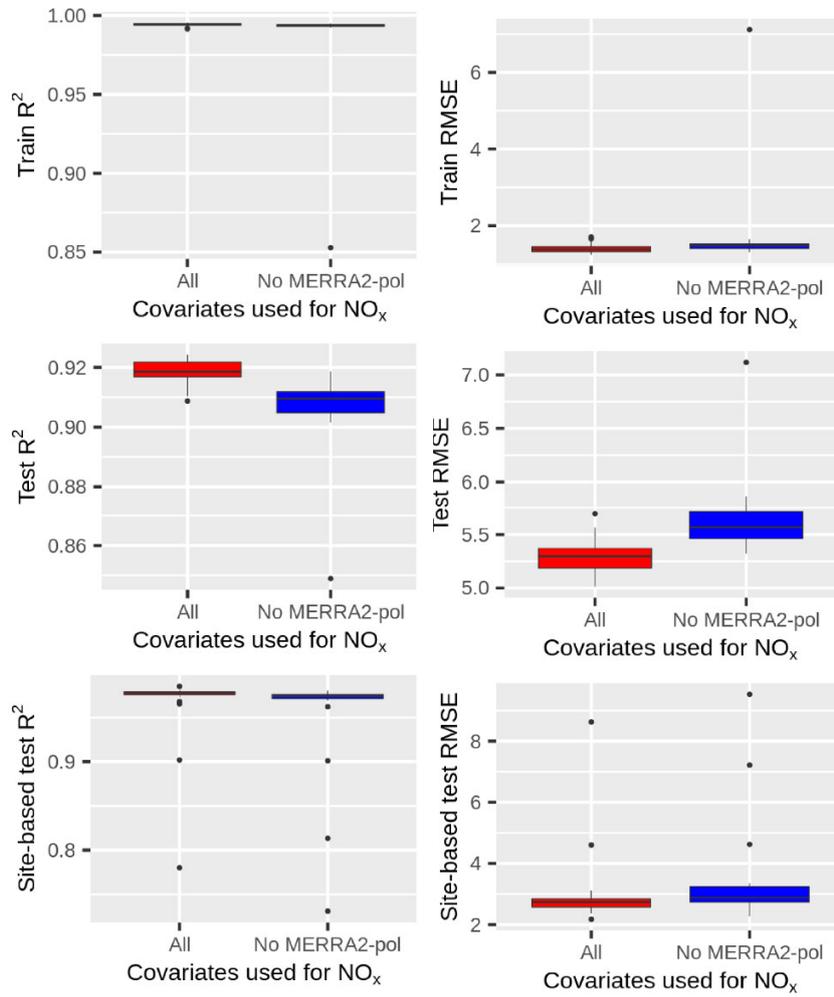

**Fig. S14**
**A comparison of California meteorology between 2019 and 2020.**

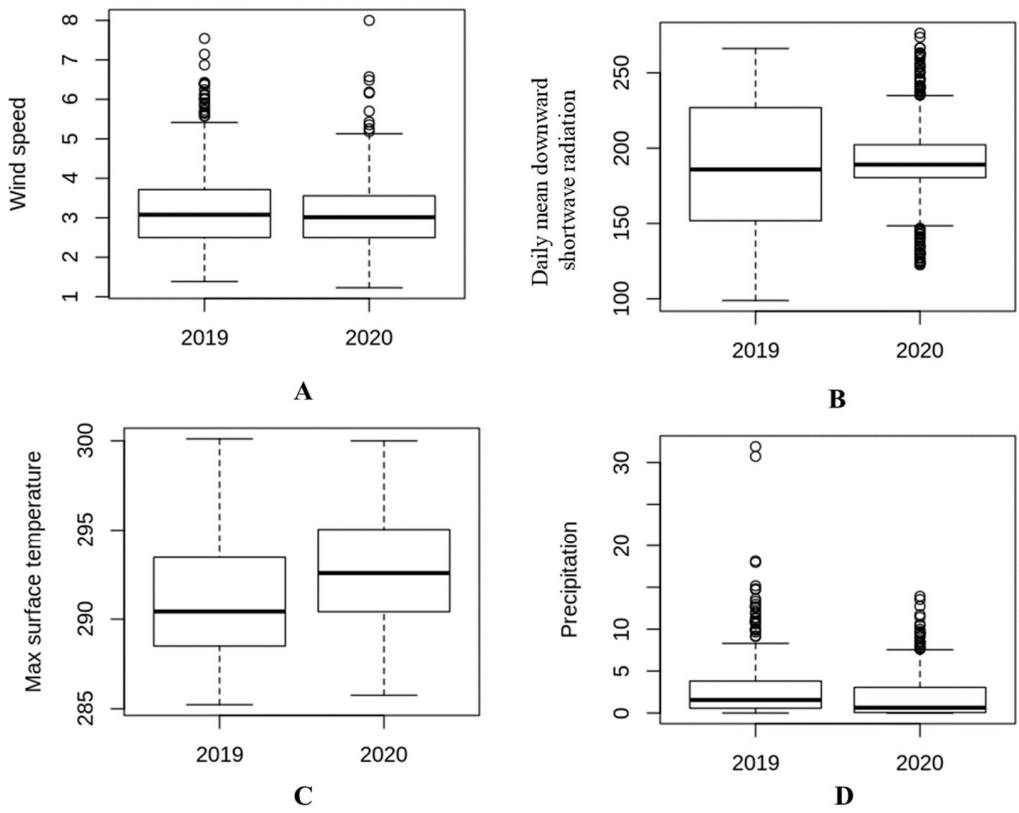

**Fig. S15**

**Temporal evolution of ensemble predicted NO$_2$ of a 2020 week by the joint PINNs for California.** A. Feb. 19 to 25, 2020, $R^2$: 0.97, RMSE: 1.00 ppb. B. Feb. 26 to March 3, 2020, $R^2$: 0.98, RMSE: 0.81 ppb. C. March 4 to 10, 2020, $R^2$: 0.98, RMSE: 0.68 ppb. D. March 11 to 17, 2020, $R^2$: 0.98, RMSE: 0.54 ppb. E. March 18 to 24, 2020, $R^2$: 0.99, RMSE: 0.46 ppb. F. March 25 to 41, 2020, $R^2$: 0.99, RMSE: 0.47 ppb

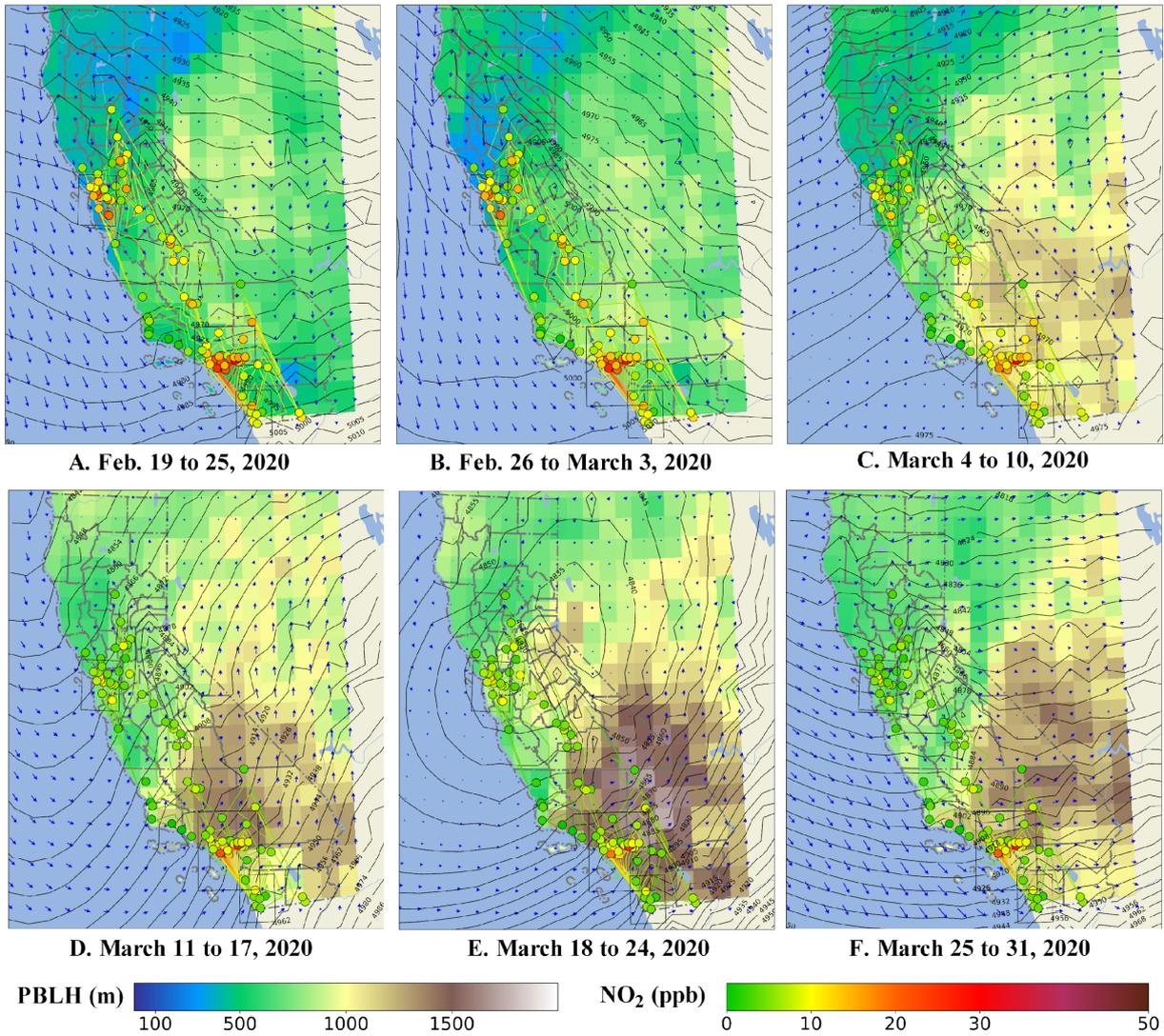

**Fig. S16**

**Temporal evolution of ensemble predicted NO$_2$ of a 2019 week by the joint PINNs for California.** A. Feb. 20 to 26, 2019, R$^2$: 0.98, RMSE: 0.92 ppb. B. Feb. 27 to March 5, 2019, R$^2$: 0.91, RMSE: 2.65 ppb. C. March 6 to 12, 2019, R$^2$: 0.98, RMSE: 0.71 ppb. D. March 13 to 19, 2019, R$^2$: 0.97, RMSE: 0.91 ppb. E. March 20 to 26, 2019, R$^2$: 0.98, RMSE: 0.74 ppb. F. March 27 to April 2, 2019, R$^2$: 0.98, RMSE: 0.79 ppb

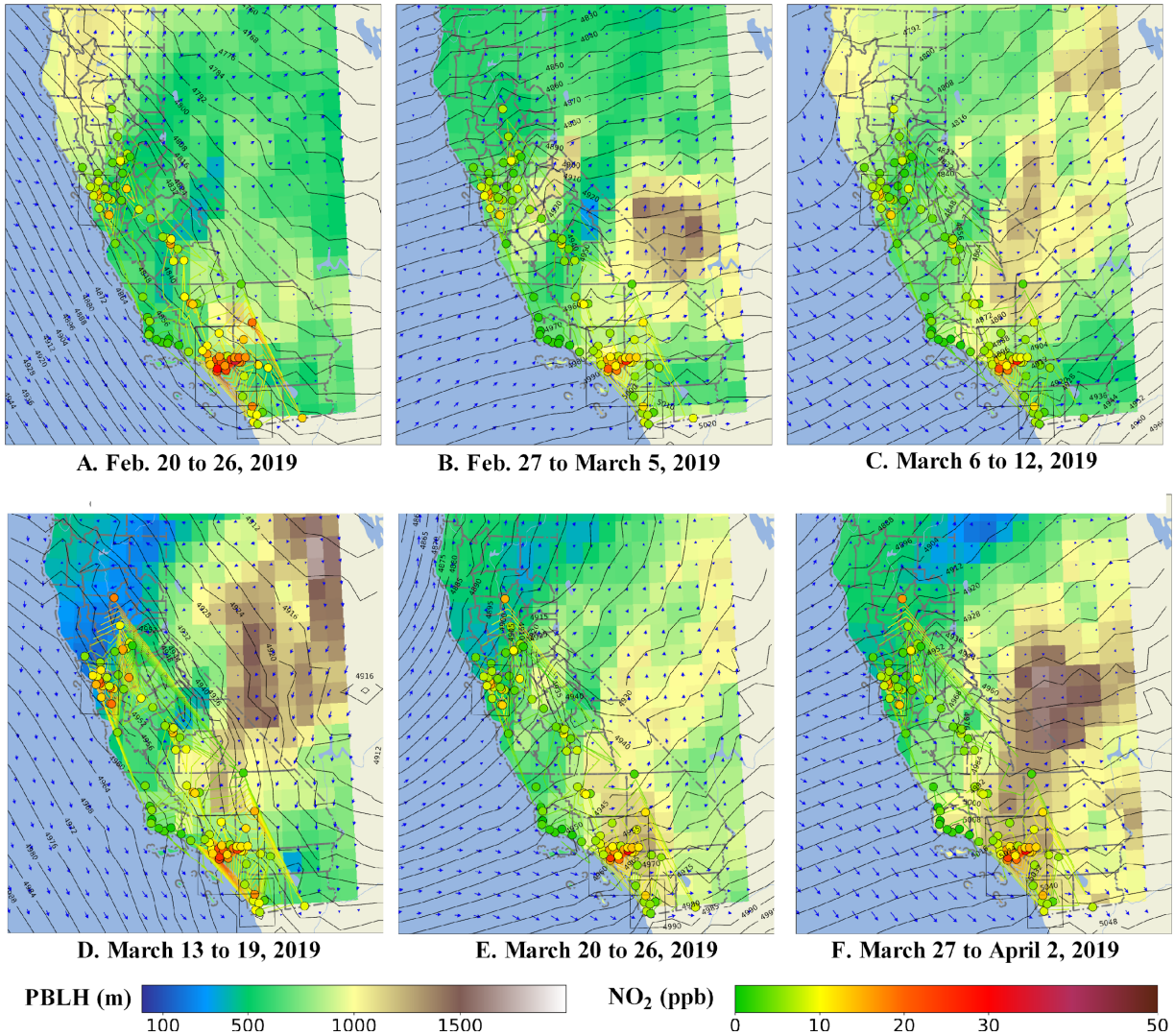

**Fig. S17**

**Temporal evolution of ensemble predicted NO₂ of a 2020 week by the joint PINNs for Los Angeles, CA.** A. Feb. 19 to 25, 2020. B. Feb. 26 to March 3, 2020. C. March 4 to 10, 2020. D. March 11 to 17, 2020. E. March 18 to 24, 2020. F. March 25 to 41, 2020.

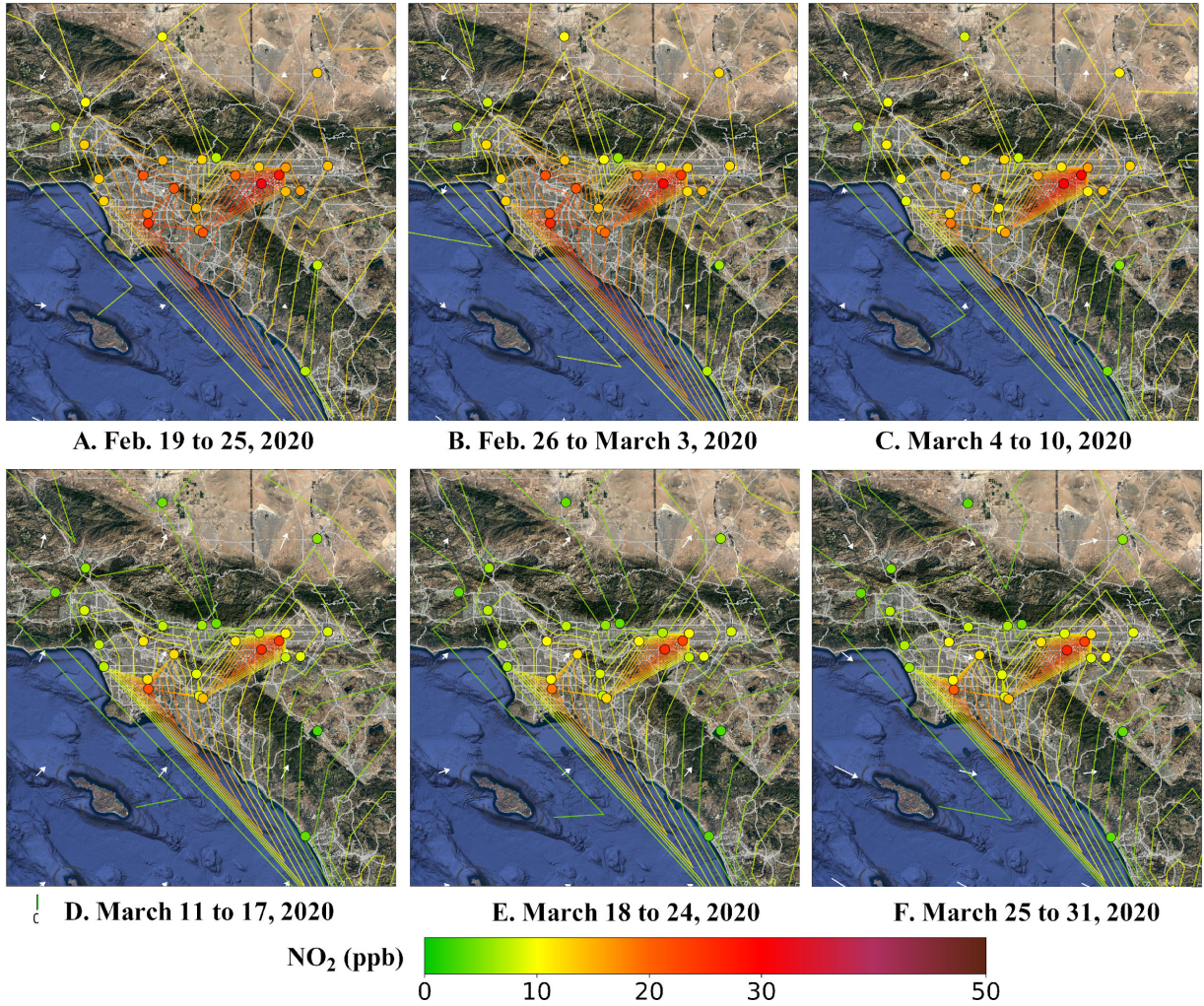

A. Feb. 19 to 25, 2020     B. Feb. 26 to March 3, 2020     C. March 4 to 10, 2020

D. March 11 to 17, 2020    E. March 18 to 24, 2020         F. March 25 to 31, 2020

NO$_2$ (ppb)

0   10   20   30   50

**Fig. S18**

**Temporal evolution of ensemble predicted NO$_2$ of a 2019 week by the joint PINNs for Los Angeles.** A. Feb. 20 to 26, 2019. B. Feb. 27 to March 5, 2019. C. March 6 to 12, 2019. D. March 13 to 19, 2019. E. March 20 to 26, 2019. F. March 27 to April 2.

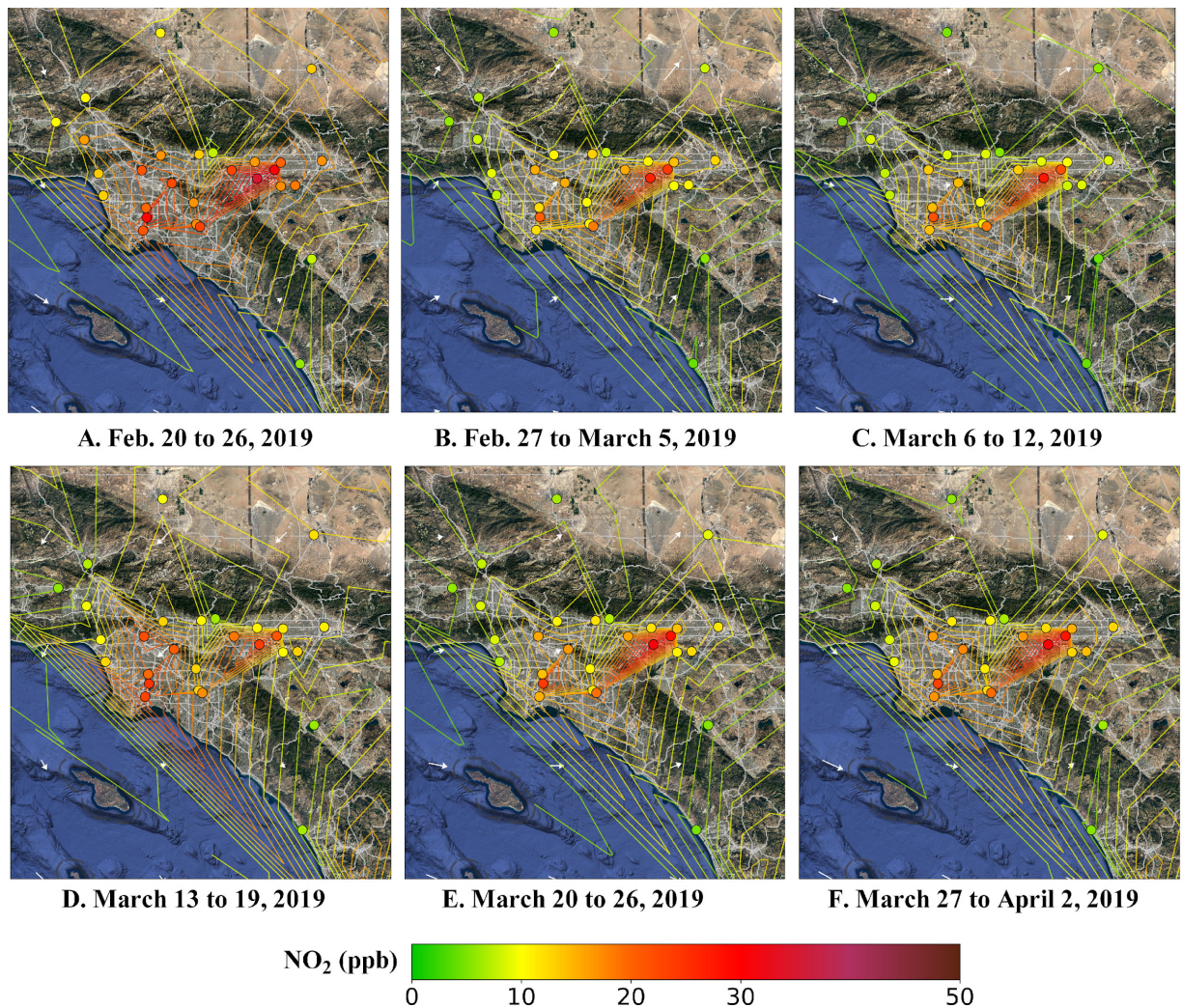

**Fig. S19**
**Feature importance for our jPINNS.** A. $NO_2$. B. $NO_x$.

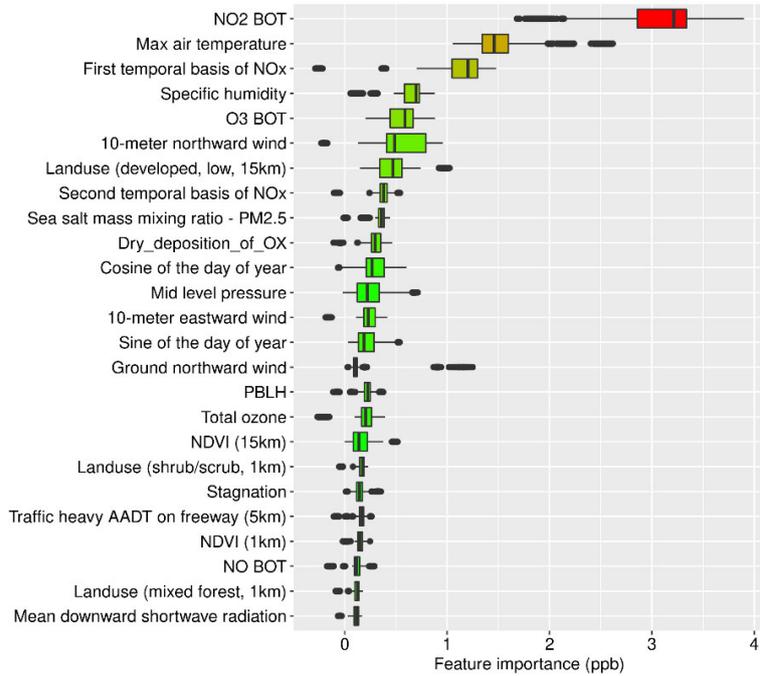

A. Feature importance of $NO_2$

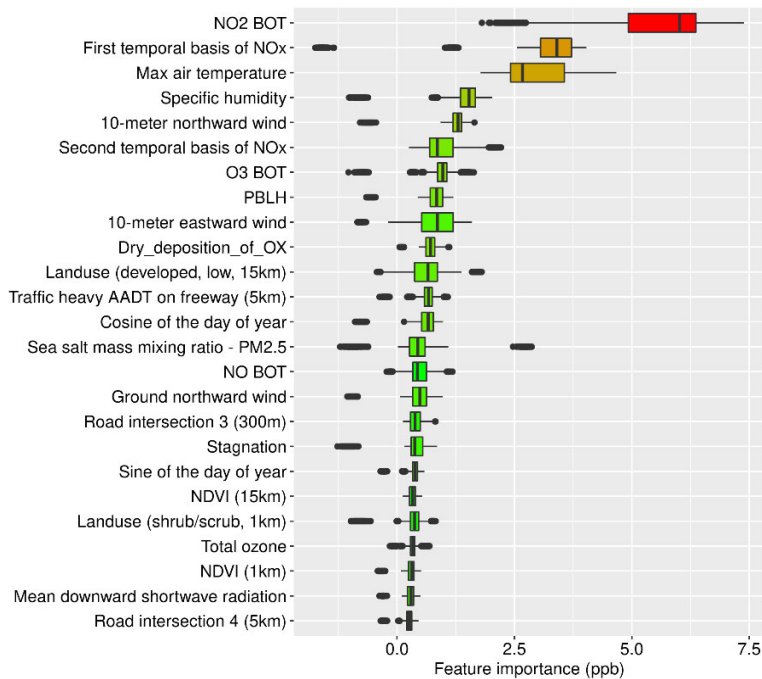

B. Feature importance of $NO_x$

**Table S1.**

Input variables and data sources.

| Class | Variable | Unit | Description and source |
|---|---|---|---|
| Coordinates | Latitude | ° | Latitude and longitude: directly from data coordinates; elevation: space shuttle radar topology mission (SRTM) |
| | Longitude | ° | |
| Elevation | Elevation | meter | |
| Time | Day of year | - | To capture temporal variability |
| | Week index | | Temporal index |
| Traffic variables | Scaled heavy duty traffic load on freeways (300m) | Emissions-scaled annual average daily vehicle meters | Traffic load (annual average daily vehicle meters) within 300 m and 5 km buffers. Traffic load was calculated from Streetlytics traffic volume estimates (2016 for Southern California and 2019 for Northern California) for vehicles on freeways and other roads. Traffic load was scaled over years prior to 2020 using a 2% year-over-year increase factor. During 2020-2021, traffic volume was significantly impacted by the coronavirus pandemic. Traffic volumes were scaled using county-level scaling factors estimated from changes in total traffic volume reported in the Caltrans Performance Measurement System PeMS system. A statewide average scaling factor was used to adjust heavy duty traffic volumes for the COVID period. Traffic load was scaled over years to reflect changes in per vehicle emissions over time separately for for heavy duty and light duty vehicles using scaling factors calculated using EMFAC2017 – v7. 2016 was used as the baseline emissions level (i.e., scaling factor = 1). |
| | Scaled heavy duty traffic load on freeways (5km) | | |
| | Scaled light duty traffic load on freeways (300m) | | |
| | Scaled light duty traffic load on freeways (5km) | | |
| | Scaled light duty traffic load on non-freeways (300m) | | |
| | Scaled heavy duty traffic load on non-freeways (5km) | | |
| | CALINE4 Simulated $NO_x$ on freeways (functional class 1 and 2 roads) and other roads (functional class 3-5 roads) | ppb | Monthly average concentration of $NO_x$ were estimated with the CALINE4 line source dispersion model, using traffic volumes assigned along roadways with vehicle emission factors from the California Air Resources Board Emissions Factor (EMFAC) model and the NOAA Real-Time Mesoscale Analysis meteorological re-analysis model. |
| | Road intersection of class 1 (300m) | count/km$^2$ | The density of intersections in 300 and 5000 meter buffers. Intersections were divided into 4 classes based on the type of roadway involved in the intersection: class 1 - Only major roadways; class 2 - Major roads intersecting minor roads; class 3- Only minor roads; and class 4 - Any road with restricted through traffic. |
| | Road intersection of class 1 (5km) | | |
| | Road intersection of class 2 (300m) | | |
| | Road intersection of class 2 (5km) | | |
| | Road intersection of class 3 (300m) | | |
| | Road intersection of class 3 (5km) | | |
| | Road intersection of class 4 (300m) | | |
| | Road intersection of class 4 (5km) | | |
| High-resolution meteorology | Min air temperature | °C | From gridMET (http://www.climatologylab.org/gridmet.html) (spatial resolution: 4km; temporal resolution: daily) (*26*). |
| | Max air temperature | °C | |
| | Mean wind speed | m/s | |
| | Mean specific humidity | g/kg | |
| | Mean downward shortwave radiation | watt/meter$^2$, | |
| | Accumulated precipitation, (millimeters of rain per meter$^2$ in 1h). | mm/m$^2$ | |
| PBLH | Planetary boundary layer height (PBLH) | meter | To represent potential for dilution of air pollutants. From NASA MINDS. |
| MINDS | Carbon monoxide | mol mol-1 | Bottom layer diagnostics (50-60 meters above the ground): Hourly values for a variety of surface-level values, including PM, |
| | Dry deposition of OX | kg m-2 s-1 | |
| | Nitric oxide | mol mol-1 | |

| | Nitrogen dioxide | mol mol-1 | PM$_{2.5}$, NO, and NO$_2$. Obtained directly from Dr. Luke Oman, NASA. |
| --- | --- | --- | --- |
| | Ozone | mol mol-1 | |
| | Black Carbon Mass Mixing Ratio | kg/kg | |
| | Nitrate Mass Mixing Ratio [PM$_{2.5}$] | kg/kg | |
| | Mid level pressure | Pa | |
| | Sea salt surface mass concentration PM $_{2.5}$ | mol mol-1 | |
| | 50-meter eastward wind | m s-1 | Single-level diagnostics: basic meteorological information such as temperature and humidity at 2m and 10m. These may be useful model inputs |
| | 10-meter eastward wind | m s-1 | |
| | 2-meter eastward wind | m s-1 | |
| | 50-meter northward wind | m s-1 | |
| | 10-meter northward wind | m s-1 | |
| | 2-meter northward wind | m s-1 | |
| | 10-meter specific humidity | kg kg-1 | |
| | 2-meter specific humidity | kg kg-1 | |
| | 10-meter air temperature | ºC | |
| | 2-meter air temperature | ºC | |
| | Surface skin temperature | ºC | |
| | Surface pressure | Pa | |
| | Total ozone | Dobsons | |
| Derived wind variables | Vertical stagnation of wind speed | m/s | $w_{stag} = \left(\sqrt{u_{50}^2 + v_{50}^2} - \sqrt{u_{10}^2 + v_{10}^2}\right)$, derived from wind speeds of single-level diagnostics. |
| | Wind shear/mechanical mixing | m/s | $w_{mix} = \left(\sqrt{u_{10}^2 + v_{10}^2} - \sqrt{u_2^2 + v_2^2}\right)$, derived from wind speeds of single-level diagnostics. |
| Land-use | Open water | % | The land cover areal proportion within the buffers of 1km and 15km was calculated based on the National Land Cover Database (NLCD) (https://www.mrlc.gov) (*24*). Average monthly NDVI from NASA's Aqua and Terra satellite (MOD13A2 V6 and MCD13A2 V6) (https://lpdaac.usgs.gov/). |
| | Developed, open space | % | |
| | Developed, low intensity | % | |
| | Developed, medium intensity | % | |
| | Developed, high intensity | % | |
| | Barren land | % | |
| | Deciduous forest | % | |
| | Evergreen forest | % | |
| | Mixed forest | % | |
| | Shrub/scrub | % | |
| | Grassland/Herbaceous | % | |
| | Pasture/Hay | % | |
| | Cultivated crops | % | |
| | Woody wetlands | % | |
| | Emergent herbaceous wetlands | | |
| | Impervious surface | % | |
| | NDVI | | |
| Food facilities | Shortest distance from fast food | m | OpenStreet |
| | Shortest distance from restaurants | m | |
| | Kernel density for fast food (30 km) | | |
| | Kernel density for restaurants (30 km) | | |
| Temporal basis function | The first and second temporal basis functions | | Extracted from AQS routine monitoring sites using iterative singular value decomposition (*21*) and represented seasonal changes for the study region. |
| Others | Shortest distance from shorelines | m | Obtained based on California shoreline. |
| | Population | people /km$^2$ | Interpolated from U.S. Census data for 2000 and 2010 and American Community Survey data for 2019 (five year average centered on 2017). Census data was scaled to yearly data by linear interpolation between each consecutive pair. Population for 2018 through 2021 was extrapolated using the equation calculated for 2010 to 2014=7 |

| | | |
|---|---|---|
| Shortest distance from airports | m | Airport emissions were estimated using emissions totals provided by CARB for baseline year 2012. Emissions were scaled to annual values for 2004-2019 using count of FAA-reported aircraft operations. For 2020-2021, emissions were scaled on a weekly basis using jet fuel refinery production reported by the California Energy Commission. The distance of each monitor location to any airport and to the nearest airport with high emissions (>0.1 tons NO2 per day). |
| Shortest distance from airports with high emissions | m | |

Note: the covariates of MINDS pollutants and land-use shown in gray background were removed in sensitivity analysis of the jPINN.

**Table S2.**

Comparison of generalization for our approach with other representative methods.

| Pol. | Model | Site-based testing | | Regular testing | | Training | |
|---|---|---|---|---|---|---|---|
| | | $R^2$ | RMSE (ppb) | $R^2$ | RMSE(ppb) | $R^2$ | RMSE(ppb) |
| $NO_2$ | Joint PINN[a] | 0.95 (0.77-0.98)[b] | 1.54 (1.19-3.26)[c] | 0.92 (0.88-0.97) | 2.16 (1.31-2.79) | 0.96 (0.88-0.98) | 1.44 (0.90-3.88) |
| | Joint PINN (no ele.)[d] | 0.94 (0.56-0.97) | 1.67 (1.35-4.81) | 0.92 (0.89-0.94) | 1.46 (1.36-2.12) | 0.96 (0.93-0.97) | 1.46 (1.36-2.12) |
| | Separate PINN[e] | 0.87 (0.61-0.94) | 2.63 (1.81-5.08) | 0.86 (0.60-0.93) | 2.74 (2.12-6.05) | 0.87 (0.61-0.94) | 2.63 (1.87-5.92) |
| | XGBoost | 0.75 (0.60-0.84) | 3.81 (2.96-4.76) | 0.92 (0.92-0.93) | 2.18 (2.08-2.31) | 0.94 (0.94-0.95) | 1.80 (1.71-1.88) |
| | Random forest | 0.65 (0.47-0.78) | 4.51 (3.43-5.60) | 0.89 (0.87-0.91) | 2.58 (2.46-2.72) | 0.97 (0.97-0.98) | 1.25 (1.20-1.32) |
| | Baseline FRNN[f] | 0.56 (-0.43-0.81) | 5.51 (3.25-14.74) | 0.92 (0.90-0.93) | 2.14 (2.06-2.28) | 0.94 (0.93-0.95) | 1.91 (1.83-2.00) |
| $NO_x$ | Integral PINN | 0.96 (0.73-0.98) | 3.25 (2.38-7.81) | 0.91 (0.87-0.97) | 5.52 (2.99-7.00) | 0.97 (0.73-0.99) | 2.89 (1.81-6.87) |
| | Integral PINN (no ele.) | 0.93 (0.27-0.96) | 4.23 (3.24-13.59) | 0.92 (0.89-0.93) | 2.17 (2.03-2.62) | 0.96 (0.92-0.97) | 3.70 (3.33-5.44) |
| | Separate PINN | 0.80 (0.62-0.83) | 4.15 (2.94-7.25) | 0.79 (0.63-0.83) | 4.29 (3.46-6.72) | 0.80 (0.62-0.83) | 4.24 (3.38-6.59) |
| | XGBoost | 0.66 (0.51-0.78) | 10.31 (7.49-12.93) | 0.91 (0.89-0.92) | 5.42 (5.04-5.96) | 0.94 (0.94-0.95) | 4.31 (3.98-4.57) |
| | Random forest | 0.60 (0.43-0.70) | 11.27 (8.23-14.06) | 0.88 (0.85-0.89) | 6.48 (6.01-7.07) | 0.97 (0.96-0.97) | 3.18 (3.00-3.37) |
| | Baseline FRNN | 0.53 (-0.44-0.79) | 14.35 (7.06-41.42) | 0.91 (0.89-0.92) | 5.43 (5.03-5.88) | 0.93 (0.92-0.94) | 4.90 (4.64-5.37) |

[a] Our joint physics-informed neural network;
[b] Mean $R^2$ ($R^2$ range: minimal, maximum) for 150 models;
[c] Mean RMSE (RMSE range: minimal, maximum) for 150 models;
[d] Joint physics-informed neural network with no elevation PDE;
[e] Separate physics-informed neural networks (not joint output) for $NO_2$ and $NO_x$ respectively.

**References and Notes**

1.  J. Pedlosky, *Geophysical fluid dynamics*. (Springer, 1987).
2.  S. Ulfah, S. A. Awalludin, Wahidin, Advection-diffusion model for the simulation of air pollution distribution from a point source emission. *J Phys Conf Ser* **948**, (2018).
3.  L. Li, Y. Fang, J. Wu, J. Wang, G. Y., Encoder-Decoder Full Residual Deep Networks for Robust Regression Prediction and Spatiotemporal Estimation. *IEEE Transactions on Neural Networks and Learning Systems* **32**, 4217-4230 (2021).
4.  D. Andrews, *An Introduction to Atmospheric Physics*. (Cambridge University, Cambridge 2010).
5.  EPA. (2021).
6.  A. Vaswani *et al.*, "Attention is all you need," (arXiv preprint arXiv:1706.03762., 2017).
7.  P. Zhou *et al.*, in *Proceedings of the 54th annual meeting of the association for computational linguistics*. (2016), vol. 2, pp. 207-212.
8.  S. Ioffe, C. Szegedy, in *ICML*. (2015).
9.  M. Abadi *et al.*, in *12th USENIX symposium on operating systems design and implementation (OSDI 16)*. (2016), pp. 265-283.
10. I. Goodfellow, Y. Bengio, A. Courville, *Deep Learning*. (MIT Press, 2016).
11. K. Mahendru. (2019).
12. S. Kumar, A. Srivastava, in *Proc. 18th ACM SIGKDD Conf. Knowl. Discovery Data Mining*. (Association for Computing Machinery, Beijing, 2012), pp. 08/12/2012-2008/2016/2012.
13. T. Hastie, R. Tibshirani, J. H. Friedman, *The Elements of Statistical Learning*. (Springer, New York, 2009).
14. J. D. Fast *et al.*, Modeling regional aerosol and aerosol precursor variability over California and its sensitivity to emissions and long-range transport during the 2010 CalNex and CARES campaigns. *Atmos Chem Phys* **14**, 10013-10060 (2014).
15. W. D. Van Vorst, Impact of the California clean air act. *International journal of hydrogen energy* **22**, 31-38 (1997).
16. C. Mullen, A. Flores, S. Grineski, T. Collins, Exploring the distributional environmental justice implications of an air quality monitoring network in Los Angeles County. *Environmental Research* **206**, 112612 (2022).
17. M. Franklin *et al.*, Predictors of intra-community variation in air quality. *J Expo Sci Environ Epidemiol* **22**, 135-147 (2012).
18. S. Fruin *et al.*, Spatial Variation in Particulate Matter Components over a Large Urban Area. *Atmos Environ (1994)* **83**, 211-219 (2014).
19. L. Li *et al.*, Constrained Mixed-Effect Models with Ensemble Learning for Prediction of Nitrogen Oxides Concentrations at High Spatiotemporal Resolution. *Environmental Science and Technology* **(in press)**, (2017).
20. L. Li *et al.*, Cluster-based bagging of constrained mixed-effects models for high spatiotemporal resolution nitrogen oxides prediction over large regions. *Environ Int* **128**, 310-323 (2019).
21. B. Finkenstadt, L. Held, V. Isham, *Statistical Methods for Spatio-Temporal Systems*. (Chapman & Hall/CRC, New York, 2007).
22. H. Chen, S. Bai, D. Eisinger, D. Niemeier, M. Claggett, Predicting near-road PM2. 5 concentrations: comparative assessment of CALINE4, CAL3QHC, and AERMOD. *Transportation research record* **2123**, 26-37 (2009).
23. B. P., "CALINE4--A dispersion model for predicting air pollutant concentrations near roadways. Prepared by the California Department of Transportation," (the California Department of Transportation, Sacramento, CA, 1989).
24. L. Yang *et al.*, A new generation of the United States National Land Cover Database: Requirements, research priorities, design, and implementation strategies. *ISPRS journal of photogrammetry and remote sensing* **146**, 108-123 (2018).
25. EPA. (2022).
26. T. J. Abatzoglou, Development of gridded surface meteorological datafor ecological applications and modelling. *International Journal of Climatology* **2011**, (2011).
27. L. N. Lamsal *et al.*, Ozone Monitoring Instrument (OMI) Aura nitrogen dioxide standard product version 4.0 with improved surface and cloud treatments. *Atmos Meas Tech* **14**, 455-479 (2021).